\documentclass[journal]{IEEEtran}
\usepackage{cite}
\usepackage{hyperref}

\usepackage{color}
\ifCLASSINFOpdf
\usepackage[pdftex]{graphicx}
\usepackage{subfigure}
\else
\fi
\usepackage{amsmath}
\usepackage{amsthm}
\usepackage{url}

\usepackage{times}
\usepackage{soul}
\usepackage{url}
\usepackage{graphicx}
\usepackage{amsmath}
\usepackage{amsthm}
\usepackage{booktabs}
\usepackage{algorithm}
\usepackage{algorithmic}
\usepackage{multirow}
\usepackage{float}  
\usepackage{amsfonts,amssymb}

\begin{document}
	%
	\title{Investigating Typed Syntactic Dependencies for Targeted Sentiment Classification Using Graph Attention Neural Network}

	
	\author{\IEEEauthorblockN{Xuefeng Bai,
			Pengbo Liu,
			Yue Zhang}
		
		\thanks{
			\textcircled{c}2020 IEEE.  Personal use of this material is permitted.  Permission from IEEE must be obtained for all other uses, in any current or future media, including reprinting/republishing this material for advertising or promotional purposes, creating new collective works, for resale or redistribution to servers or lists, or reuse of any copyrighted component of this work in other works.
			
			This work has been supported by National Natural Science Foundation of China under grant No.61976180. (\textit{Corresponding author: Yue Zhang})
			
			Xuefeng Bai is with the Zhejiang University, Hangzhou 310007, China, and also with the School of Engineering, Westlake University, Hangzhou 310024, China (email: xfbai.hk@gmail.com).
			
			Pengbo Liu is with the Machine Intelligence and Translation Laboratory, School of Computer Science of Technology, Harbin Institute of Technology, Harbin 150001, China (email: liupengbo.work@gmail.com)
			
			Yue Zhang is with the School of Engineering, Westlake University, and also with the Institute of Advanced Technology, Westlake Institute for Advanced Study, Hangzhou 310024, China (email: yue.zhang@wias.org.cn).
	}}

	\markboth{IEEE/ACM TRANSACTIONS ON AUDIO, SPEECH, AND LANGUAGE PROCESSING}%
	{Bai \MakeLowercase{\textit{et al.}}: Exploiting Typed Syntactic Dependencies for Targeted Sentiment Classification Using Graph Attention Neural Network}
	%



	\IEEEtitleabstractindextext{%
		\begin{abstract}
			Targeted sentiment classification predicts the sentiment polarity on given target mentions in input texts.
			Dominant methods employ neural networks for encoding the input sentence and extracting relations between target mentions and their contexts.
			Recently, graph neural network has been investigated for integrating dependency syntax for the task, achieving the state-of-the-art results.
			However, existing methods do not consider dependency label information, which can be intuitively useful.
			To solve the problem, we investigate a novel relational graph attention network that integrates typed syntactic dependency information.
			Results on standard benchmarks show that our method can effectively leverage label information for improving targeted sentiment classification performances.
			Our final model significantly outperforms state-of-the-art syntax-based approaches.
		\end{abstract}
		
		}
	
	\maketitle

	\IEEEdisplaynontitleabstractindextext

	%
	\IEEEpeerreviewmaketitle

	\section{Introduction}
	%
	%
	%
	%
	\IEEEPARstart{T}{argeted} sentiment classification~\cite{jiang-etal-2011-target,dong-etal-2014-adaptive,Vo2015TargetDependentTS,Zhang2016GatedNN,wang-etal-2018-target} is the task of predicting the sentiment polarity for target entity mentions in a given sentence. 
	For example, suppose that a sentence is ``\textit{I like
		the food here, but the service is terrible.}'' and the given targets are ``\textit{food}'' and ``\textit{service}''. 
	The output sentiment polarities on the two targets are \textit{positive} and \textit{negative}, respectively. 
	Different from text level sentiment classification~\cite{pang-etal-2002-thumbs,Meena07,yessenalina-etal-2010-multi}, targeted-sentiment is entity-centric, and therefore can offer more fine-grained opinion information from text documents.
	Dominant methods for targeted sentiment employ neural network models to encode the input sentence and target mention~\cite{wang-etal-2016-attention,Wang2018TargetSensitiveMN}. 
	Gates~\cite{Vo2015TargetDependentTS,tang-etal-2016-effective,Zhang2016GatedNN}, convolution~\cite{li-etal-2018-transformation,huang-carley-2018-parameterized}, attention~\cite{Ma2017InteractiveAN,tang-etal-2019-progressive} and memory network~\cite{Tang2016AspectLS,Chen2017RecurrentAN,Wang2018TargetSensitiveMN} have been exploited to capture the relation between the target mention and its context information. 
	The assumption is that deep syntactic and semantic features can be represented by neural encoding. 
	
	With the advance of structured neural encoders~\cite{dong-etal-2014-adaptive,tai-etal-2015-improved,kipf2017semi,velickovic2018gat}, syntactic structures predicted by external parsers have shown their usefulness for the task.
	Intuitively, syntactic structures such as dependency trees can help better encode the correlation between a target mention and the relevant sentiment keywords.
	Recent methods consider dependency trees as adjacency matrices, using graph neural networks such as graph convolutional networks (GCN~\cite{kipf2017semi}) and graph attention network (GAT~\cite{velickovic2018gat}) to encode the input sentence according to such matrices~\cite{sun-etal-2019-aspect,huang-carley-2019-syntax,zhang-etal-2019-aspect}. 
	\begin{figure}[t!]
		\centering 
		\subfigure[]{\includegraphics[width=0.95\hsize,height=0.15\hsize]{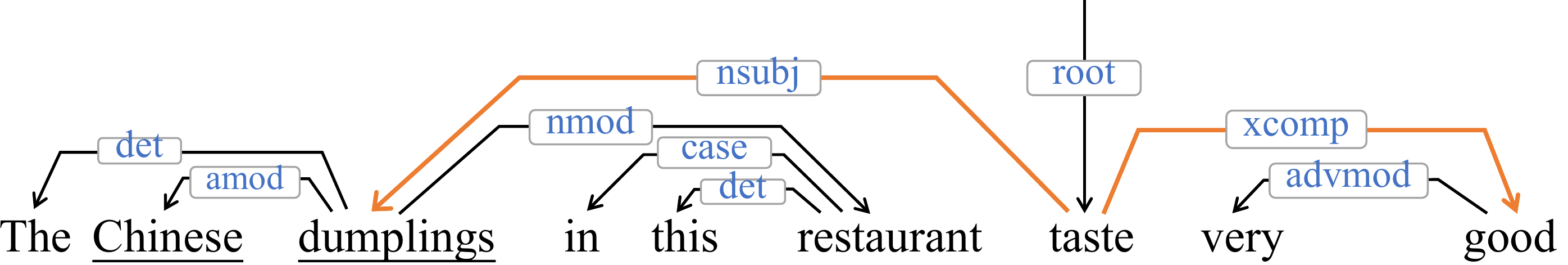}\label{fig:1-1}}\\ \hspace{-1pt}        
		\subfigure[]{\includegraphics[width=0.95\hsize]{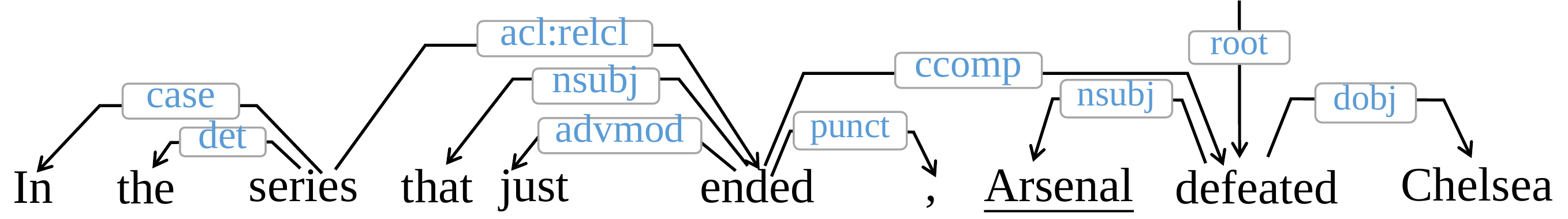}\label{fig:1-2}} \\
		\subfigure[]{\includegraphics[width=0.95\hsize]{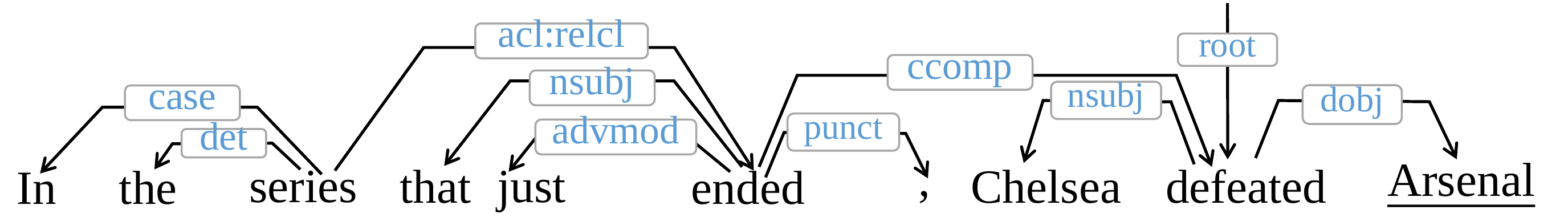}\label{fig:1-3}} \\ 
		\caption{(a) An example sentence with a dependency tree, (b,c) Two sentences which have similar dependency trees. The target mentions are underlined.}
		\label{fig:example}
	\end{figure}
	As shown in Figure~\ref{fig:1-1}, such dependency tree structures can help bring target mentions closer to its relevant contexts, thereby facilitating feature representation. 
	In this example, the relevant sentiment word ``good'' is distant from the target mention ``Chinese dumplings'' in the surface string, but close in the dependency tree (i.e., ``Chinese dumplings''
	$\stackrel{nsubj}{\longleftarrow}$``taste''$\stackrel{xcomp}{\longrightarrow}$``good'').
	
	While being more effective compared with the state-of-the-art approaches that do not use syntactic structures, these methods do not consider dependency labels, which can potentially be useful for sentiment disambiguation. 
	As shown in Figure~\ref{fig:1-2} and Figure~\ref{fig:1-3}, while ``Arsenal'' has similar dependency arc relations with the word ``defeated'' in both sentence, the sentiment polarities  are different.
	Apparently differentiating the label types can help differentiate such cases.
	On the other hand, considering arc label information can make the syntactic structure more sparse and thus increase the difficulty in learning. 
	Thus it remains a open research question how to effectively make use of such fine-grained syntax features for better targeted sentiment classification.
	
	We investigate a graph attention network to integrate such typed dependency features. 
	In particular, the proposed model incorporates label features into the attention mechanism, using a novel extended attention function to guide information propagation from a target mention's syntactic context to target mention itself. 
	We name our model relational graph attention network (RGAT).
	With the help of these label features,  our model can better capture the relationship between words, thus addressing the problems shown in Figure~\ref{fig:example}.
	Moreover, the dependency labels can serve as additional features, enriching the representation of words. 
	
	Experiments over four benchmarks show that using GAT to encode the input gives better results compared to a Transformer encoder, which coincides with recent observation that syntax is useful for targeted sentiment classifications~\cite{sun-etal-2019-aspect,huang-carley-2019-syntax,zhang-etal-2019-aspect}. 
	Further, adding typed dependency features gives consistently better results compared with a baseline without such information, thereby proving our research hypothesis. 
	{
		Our model gives significantly better results than  state-of-the-art syntax-based work on the standard Laptop, Restaurant, Twitter and MAMS datasets.
	} 
	To our knowledge, we obtain the best reported results on all datasets without using external resources.
	Our code is released at \url{https://github.com/muyeby/RGAT-ABSA}.
	
	\section{Related Work}
	With regard to structures, existing work on targeted sentiment classification can be classified into two main categories, namely those methods that do not rely on external syntax information and those using syntax structures.
	\subsection{Conventional Methods}
	Most work along the first line splits a given input sentence into two sections, including the target mention and its context. Features are extracted from each section and combining the features for making prediction. 
	For example, Vo and Zhang~\cite{Vo2015TargetDependentTS} use word2vec embeddings and pooling mechanisms for extracting features from target mention and its left and right context, respectively, before concatenating all the features for classification.
	Zhang~\textit{et al.}~\cite{Zhang2016GatedNN} use gated recurrent neural network for extracting features from target mentions and its context, before further defining a gate to integrate such features.
	Tang~\textit{et al.}~\cite{tang-etal-2016-effective} exploit two long-short memory network (LSTM~\cite{HochSchm97}) for feature encoding, and combine the last hidden state of two networks for classification.
	
	There are also attempts exploiting convolutional neural networks (CNN) for targeted sentiment classification. 
	For instance, 
	Huang and Carley~\cite{huang-carley-2018-parameterized} use parameterized filters and parameterized gates to incorporate aspect information into CNN, and apply the resulting CNN to encode the sentence.
	Xue and Li~\cite{xue-li-2018-aspect} further employ a gated CNN layer to extract aspect-specific features from the hidden states originated from a bi-directional RNN layer.
	
	The attention mechanism~\cite{Bahdanau2015NeuralMT,Vaswani2017AttentionIA} is also shown to be useful for the task.
	Ma~\textit{et al.}~\cite{Ma2017InteractiveAN} use a bidirectional attention network, which learns attention weights from the target mention to its context vectors, to model the target-context relationship. 
	Li~\textit{et al.}~\cite{li-etal-2018-hierarchical} further improve the attention-based models with position information.
	Tang~\textit{et al.}~\cite{tang-etal-2019-progressive} introduce a self-supervised attention learning approach, which automatically mines useful supervision information to refine attention mechanisms.
	
	In addition to attention, Memory networks have also been applied to this task. 
	Tang~\textit{et al.}~\cite{Tang2016AspectLS} develop a deep memory network (MemNet), which uses pre-trained word vectors as a memory and exploits attention mechanism to update the memory.
	Chen~\textit{et al.}~\cite{Chen2017RecurrentAN} improve MemNet by taking the hidden states generated by LSTM as memory and adopting gated recurrent units (GRU) to update the representation of target mentions.
	Wang~\textit{et al.}~\cite{Wang2018TargetSensitiveMN} deploy a targeted sensitive memory network for better information integration.
	
	Recently, contextualized language models such as BERT~\cite{devlin-etal-2019-bert}, GPT~\cite{Radford2018ImprovingLU} and ALBERT~\cite{Lan2020ALBERTAL} ELECTRA~\cite{Clark2020ELECTRA} have shown their usefulness for a wide range of natural language inference (NLI) works.
	With regard to the task of targeted sentiment classification, Song~\textit{et al.}~\cite{Song2019AttentionalEN} use BERT to encode a target mention and  it context, before applying attention to draw semantic interaction between targets and context words.
	Gao~\textit{et al.}~\cite{Gao19bertABSA} further introduce several variants to apply BERT for targeted sentiment classification, showing that incorporating target mention is helpful for BERT based models.
	Sun~\textit{et al.}~\cite{sun-etal-2019-utilizing} propose to construct an auxiliary sentence from the aspect and convert aspect-based sentiment classification into a
	sentence-pair classification task. 
	Xu~\textit{et al.}~\cite{xu-etal-2019-bert} re-train a BERT model on a customer reviews dataset and use the resulting model for targeted sentiment classification.
	Li \textit{et al.}~\cite{li-etal-2019-exploiting} explore BERT for end-to-end targeted sentiment classification, jointly learning to predict the target mentions and its sentiment polarity.
	
	It has been shown that BERT contains implicit syntactic information, which can be useful for downstream tasks~\cite{Goldberg2019AssessingBS,clark2019what}. In contrast to these efforts, however, we make explicit use of syntactic information by encoding discrete structures. Our method is orthogonal to the implicit use of knowledge in BERT, and can be combined with contextualized embeddings.
	
	\subsection{Syntax-based Methods}
	Among work that uses syntax structures, early work~\cite{Qiu2011OpinionWE,Liu2013OpinionTE} rely on manually-defined syntactic rules and use non-neural models.
	Subsequently, neural network models are explored for this task.
	For example, 
	Dong~\textit{et al.}~\cite{dong-etal-2014-adaptive} use an adaptive recursive neural network (AdaRNN) to encode a syntax tree, transformed by placing the target mention onto the root node. 
	Nguyen and Shirai~\cite{Nguyen2015PhraseRNNPR} further extend AdaRNN into a phrase-level recursive neural network (Phrase AdaRNN), which takes both dependency and constituent trees as input.
	He~\textit{et al.}~\cite{he-etal-2018-effective} use the distance on a dependency tree to guide the attention mechanism, thus helping model to focus on more important context words.
	
	Recent work uses graph neural networks to encode the syntactic structure, obtaining better results. 
	In particular, Sun~\textit{et al.}~\cite{sun-etal-2019-aspect} explore a GCN for encoding syntactical features to help the information exchange between target mention and context. 
	Similar to that, Zhang~\textit{et al.}~\cite{zhang-etal-2019-aspect} introduce a aspect-specific GCN to incorporate the syntactical information and long-range word dependencies into the classification model.
	Huang and Carley~\cite{huang-carley-2019-syntax} apply a GAT to model a dependency tree and integrate the GAT layer into LSTM to model the cross-layer dependency.
	
	\begin{figure*}[!t]
		\centering
		\includegraphics[width=0.7\textwidth]{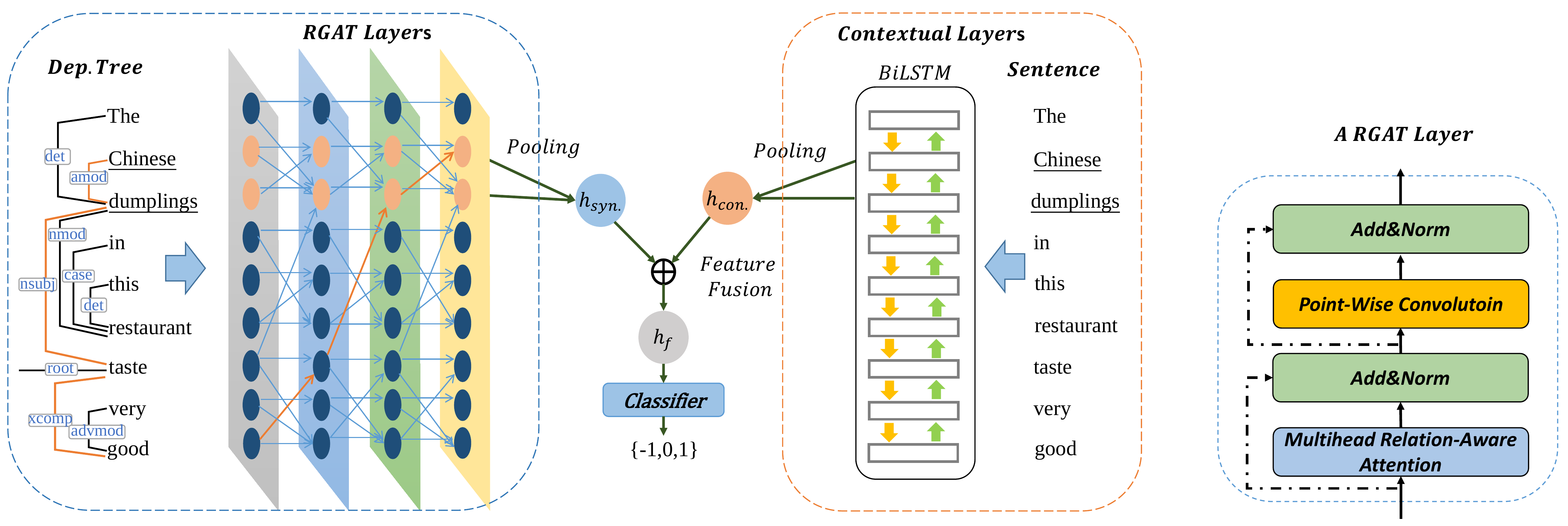}
		\caption{An overview of the proposed network for targeted sentiment classification\protect\footnotemark. It consists of a contextual encoder (orange dotted frame), a RGAT encoder (blue dotted frame) and a classifier. The BiLSTM module can be replaced by BERT.}
		\label{fig:one}
	\end{figure*}
	\footnotetext{{In order to prevent the input graph from being too sparse, we remove the edge directions and add a self-loop edge for each word.}}
	Our work is in this line. We take GAT as our base model and investigate a relational extension. 
	{
	Different from existing syntax-based work, we 1) exploit dependency relation information, which is proven to be useful by our work yet not investigated by ~\cite{sun-etal-2019-aspect,huang-carley-2019-syntax, zhang-etal-2019-aspect};
	2) use an independent encoder for structure modeling instead of exploiting syntactic structures to refine contextual representations~\cite{sun-etal-2019-aspect,huang-carley-2019-syntax,zhang-etal-2019-aspect,wang-etal-2020-relational}. Our double-encoder structure has two main advantages: a) the structure encoder is detachable and can be relatively easily applied to a new sequence encoder model (especially when the sentence length is not equal to the number of nodes in graph); b) it can reduce error propagation from automatic parsed dependency trees, as loss do not back-propagate from the tree representation to the sequence encoder.
	}

	{
		Another similar attempt to this end is Shaw \textit{et al.}~\cite{shaw-etal-2018-self}, who extends a self-attention network (SAN) by integrating relative position information for neural machine translation.
		However, the relative position information is a simpler type of relation and thus is less informative.
	} 
	In contrast to their work, we investigate relational GAT for targeted sentiment classification, with significantly improved results.

	\section{Approach}
	For our model, each training instance consists of three components: a target mention, a sentence and a dependency tree of the sentence.
	Formally, we denote these components as a triplet: $\left\langle \mathcal{T}, \mathcal{S}, \mathcal{G}\right\rangle$, 
	where $\mathcal{T}=\{w_i,w_{i+1},...,w_{i+m-1}\}$ denotes a target mention word sequence, and $\mathcal{S} = \{w_1, w_2,...w_i,...,w_{i+m},...w_n \}$ denotes a sentence. The lengths of $\mathcal{T}$ and $\mathcal{S}$ are $m$ and $n$, respectively.
	$\mathcal{G}=(\mathcal{V}, \mathcal{A}, \mathcal{R})$ represents a syntactic graph (for example, a dependency tree) over $S$, where $\mathcal{V}$ includes all nodes (or words) $\{w_1,w_2,...,w_n\}$, $\mathcal{A}$ is an adjacent matrix $\mathcal{A} \in \mathbb{R}^{n\times n}$ with $\mathcal{A}_{ij}=1$ if there is a dependency arc relation between word $w_i$ and $w_j$, and $\mathcal{A}_{ij}=0$ otherwise, and $\mathcal{R}$ is a label matrix, where $\mathcal{R}_{ij}$ records the corresponding label of $\mathcal{A}_{ij}$ if $\mathcal{A}_{ij}=1$, and $\mathcal{R}_{ij}=$ \textit{None} otherwise.
	The goal of targeted sentiment classification is to predict the sentiment polarity $y \in \{ 1, -1, 0\}$ of the sentence ${S}$ over the target mention $T$, where $1$, $-1$ and $0$ denote \textit{positive}, \textit{negative} and \textit{neutral}, respectively.
	
	The overall architecture of our model is shown in Figure~\ref{fig:one}. It is mainly composed of three components: the Contextual Encoder, the Relational Graph Attention (RGAT) Encoder and the Classifier. 
	The Contextual Encoder can be viewed as a conventional model which applies Contextual encoder layers (e.g., BiLSTM, CNN, BERT) for feature learning and uses a simple pooling function for feature aggregation.
	The RGAT encoder is a syntax based encoder which incorporates syntax information into the progress of sentence modeling, thus generates syntax-aware word embeddings.
	The Feature Fusion mechanism is designed to dynamically combine the contextual and syntactic representations (denoted as $h_{syn.}$ and $h_{con.}$).
	The final representation $h_f$ is then fed into a Classifier for classification.
	
	\subsection{Contextual Encoder}
	\label{sec:contextual}
	We consider two types of models for contextual modeling: the first is a BiLSTM model, which is widely used for targeted sentiment classification and other tasks~\cite{Sundermeyer2012LSTMNN,Bahdanau2015NeuralMT}. The other is BERT~\cite{devlin-etal-2019-bert}, which is pre-trained on large-scale raw texts and has more parameters than BiLSTM.
	
	\noindent\textbf{BiLSTM} 
	We use BiLSTM to model the bidirectional context information. Following previous work~\cite{sun-etal-2019-aspect}, we use GloVe embeddings $v_i \in \mathbb{R}^{d_e}$, POS-tag embeddings $t_i \in \mathbb{R}^{d_t}$ as well as position embeddings $p_i \in \mathbb{R}^{d_p}$ as inputs features, where
	$d_e$, $d_t$ and $d_p$ denote the dimensions of word, POS-tag and position embeddings, respectively.
	Therefore, the representation of a word $w_i$ is denoted as $x_i=[v_i;t_i;p_i]$, which is the concatenation of $v_i$, $t_i$ and $p_i$.
	
	Given an word embedding sequence $x= \{ x_1, x_2,...,x_n \}$, a forward $\overrightarrow{\text{LSTM}}$ generates a set of hidden states $\overrightarrow{h}= \{\overrightarrow{h_1}, \overrightarrow{h_2},..., \overrightarrow{h_n}\}$, and a backward $\overleftarrow{\text{LSTM}}$ generates a set of hidden states $\overleftarrow{h} = \{\overleftarrow{h_1}, \overleftarrow{h_2},..., \overleftarrow{h_n}\}$. 
	Finally, the  output hidden states $h = \{{h_1}, {h_2},..., {h_n}\}$ are obtained by concatenating the corresponding forward and backward hidden states:
	\begin{equation}
	\begin{split}
	\overrightarrow{h} &= \overrightarrow{\text{LSTM}}([x_1,x_2,...,x_n])\\
	\overleftarrow{h} &= \overleftarrow{\text{LSTM}}([x_1,x_2,...,x_n])\\
	h &= [\overrightarrow{h};\overleftarrow{h}]
	\end{split}
	\label{bilstm}
	\end{equation}
	
	\noindent\textbf{BERT} is a pre-trained masked language model, which is based on a Transformer~\cite{Vaswani2017AttentionIA}.
	Previous work have shown that BERT can significantly boost the classification accuracy~\cite{Song2019AttentionalEN,Gao19bertABSA,li-etal-2019-exploiting}. 
	For comparability, we also adopt BERT to generate contextual embeddings.
	
	To facilitate fine-tuning of the BERT model, we follow~\cite{Song2019AttentionalEN} to refactor the sequence as ``[CLS]'' + sentence + ``[SEP]'' + target mention + ``[SEP]'' as input to BERT. 
	Formally, denoting the resulting sequence as\footnote{For simplicity, we assume that the BERT tokenizer~\cite{google16} does not change the original tokenize results.} 
	\begin{equation}
	x = \{ x_0, x_1, ..., x_n, x_{n+1}, x_{n+2}, ..., x_{n+1+m}, x_{n+2+m}\},
	\end{equation}
	where $x_0$ and $x_{n+1}$ are vector representations of ``[CLS]'' and ``[SEP]'' respectively. 
	The BERT model generates a new sequence with the same length as $x$, denoted as\footnote{Please refer to~\cite{devlin-etal-2019-bert} for detailed BERT.}: 
	\begin{equation}
	h = \{ h_0, h_1, ..., h_n, h_{n+1}, h_{n+2}, ..., h_{n+1+m},h_{n+2+m}\},
	\end{equation}
	where $h_0$ is called a ``BERT pooling'' vector, $h_1, h_2, ..., h_n$ are output contextual representations of the input word sequence. 
	and $h_{n+2},...,h_{n+1+m}$ corresponds to the output embedding of target mention.
	In this work, we use $h_1, h_2, ..., h_n$ for pooling and feature fusion.
	
	Different from BiLSTM, the BERT based model does not take GloVe, POS-tag or position embeddings as additional input features, as such features are inherently learned by BERT itself\footnote{We also tried to use such features, without observing significant improvements.}.
	
	\subsection{Relational Graph Attention Network}
	\label{sec:rgat}
	A relational graph attention network (RGAT) aims to perform information exchange among words according to the syntactic dependency paths. 
	Compared with standard  GAT networks, the RGAT method can additionally make use of the labeled relations (i.e., typed syntactic dependencies), thus generating more informative representations.
	In this section, we start by introducing the baseline GAT model which operates on an unlabeled graph $\mathcal{G}=(\mathcal{V}, \mathcal{A})$, and then present RGAT, which extends the GAT with the ability to model a labeled graph $\mathcal{G} = (\mathcal{V}, \mathcal{A}, \mathcal{R})$.
	
	\subsubsection{Vanilla Graph Attention Network}
	We take the graph attention network as a baseline.\footnote{We do not take the TD-GAT model of Huang and Carley~\cite{huang-carley-2019-syntax} as our baseline, as TD-GAT contains a LSTM module which makes model more complicated and harder to analyze. 
		Besides, our preliminary experiments showed that our GAT baseline gives comparable results with TD-GAT.}
	The graph attention network~\cite{velickovic2018gat} is a variant of graph neural networks, which leverages masked self-attention layers to encode graph structures. 
	Compared with other types of graph neural networks (e.g. graph convolutional networks~\cite{kipf2017semi}, graph recurrent networks~\cite{song-etal-2018-graph}), 
	GAT can be better interpreted thanks to the attention mechanism.
	
	\noindent\textbf{Input and output} 
	A GAT takes a set of word (or node) embeddings $\{x_1, x_2,...,x_n\}$ as initial hidden states $\{h_1^0,h_2^0,...,h_n^0\}$, iteratively producing more abstract features $\{h_1^l,h_2^l,...,h_n^l\}$ with increasing $l$, $l \in [1, L]$.
	The $l$th GAT layer takes predecessor word features $\{h^{l-1}_1,h^{l-1}_2,...,h^{l-1}_n\}$ and an adjacent matrix $\mathcal{A}$ as input,
	and produces a new set of word features $\{h^l_1, h^l_2,...,h^l_n \}$ as its output.
	
	\noindent\textbf{Feature aggregation} 
	{Given a word $w_i$ with its neighbor word index $j\in \mathcal{N}(i)$\footnote{{$\mathcal{N}{(i)}$ is induced from the adjacent matrix $\mathcal{A}$.}}}, a GAT updates the word's representation at layer $l$ by calculating a weighted sum of the neighbor states. 
	Briefly, the aggregation process of a multi-head-attention-based GAT can be described as:
	\begin{equation}
	\begin{split}
	& {h}^{l}_i = \mathop{||}_{z=1}^{Z} \sigma\big(\sum_{j \in \mathcal{N}(i)} \alpha_{ij}^{lz} W^{lz}_V{h}^{l-1}_{j}\big), \\
	\end{split}
	\end{equation}
	where $||$ denotes vector concatenation, $W^{lz}_V \in \mathbb{R}^{\frac{d}{Z}\times d}$ is a parameter matrix of the $z$th head at layer $l$, $d$ denotes the dimension of word feature vectors, $Z$ is the number of attention heads,  and $\sigma$ represents the sigmoid activation function.
	The weight $\alpha_{ij}^{lz}$ models to what extent $h^{l}_i$ depends on $h^{l-1}_j$:
	\begin{equation}
	\begin{split}
	&\alpha^{lz}_{ij} = \frac{\exp(f(h^{l-1}_{i}, h^{l-1}_{j}))}{\sum_{j' \in \mathcal{N}(i)} \exp{(f(h^{l-1}_{i}, h^{l-1}_{j'})})}. \\
	\end{split}
	\label{eq:att}
	\end{equation}

	{If $j \notin \mathcal{N}(i)$, $\alpha^{lz}_{ij}=0$}. $f$ is an attention function. We use the scaled dot-product attention function~\cite{Vaswani2017AttentionIA}\footnote{We also tried other attention functions, where no significant improvement founded.}:
	\begin{equation}
	\begin{split}
	f(h^{l-1}_{i}, h^{l-1}_{j}) = \frac{(W^{lz}_Qh^{l-1}_i)^T(W^{lz}_Kh^{l-1}_j)}{\sqrt{d/Z}},
	\end{split}
	\label{eq:dotattention}
	\end{equation}
	where $W^{lz}_Q, W^{lz}_K \in \mathbb{R}^{\frac{d}{Z} \times d}$ are parameter matrices of the $z$th head at layer $l$.
	
	Apart from the aforementioned attention layer, we further add a point-wise convolution transformation (PCT) layer following the attention layer, which gives each node more information capacity. 
	The convolution kernel size is 1, and convolution is applied to every single token belonging to the input. 
	Given an output hidden states $h^l=\{h_1^l,h_2^l,...,h_n^l\}$ of the $l$th attention layer, the PCT layer is defined as:
	\begin{equation}
	PCT(h^l) = \delta(h^l\star W_{P_1}+b_{P_1})\star W_{P_2}+b_{P_2},
	\label{eq:ppff}
	\end{equation}
	where $\delta$ refers to the ReLU activation function, $\star$ denotes the convolution operation, $ W_{P_1},  W_{P_2}, b_{P_1}, b_{P_2}$ are weights and bias of two convolutional kernels.
	\subsubsection{Relational Graph Attention Network}
\begin{figure}[!t]
		\centering
		\includegraphics[width=0.45\textwidth]{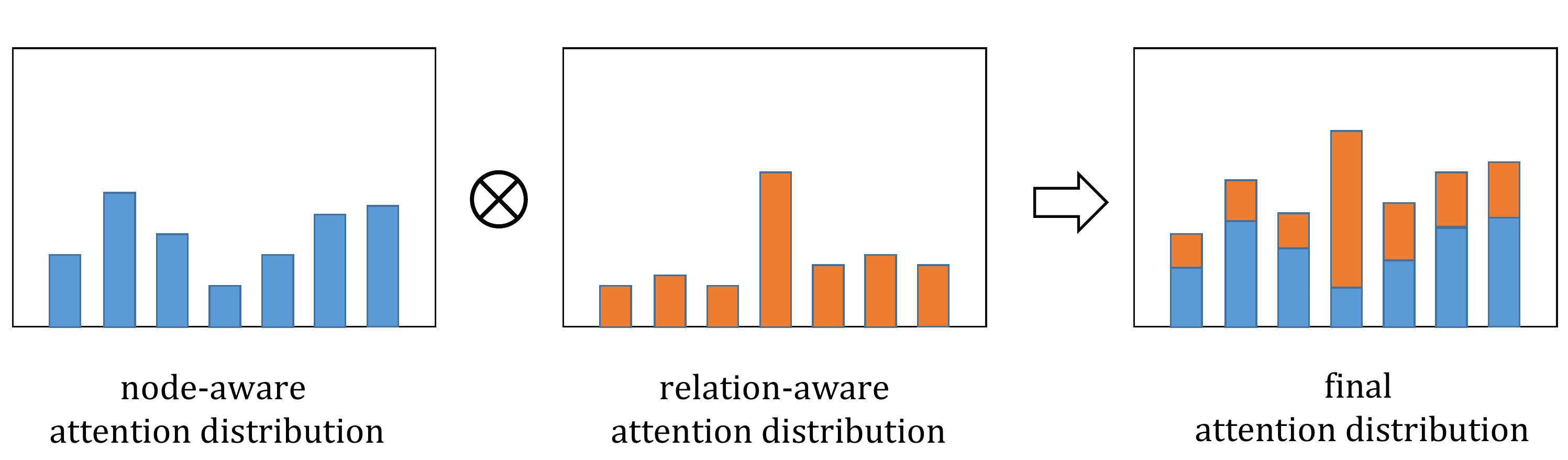}
		\caption{The attention distribution of RGAT model is a mixture of node-aware attention and relation-aware attention distribution. $\bigotimes$ denotes the attention mixing operation (see Equation~\ref{eq:ratt3}).}
		\label{fig:rgat}
	\end{figure}

	The vanilla GAT mentioned above uses an adjacent matrix as structural information, thus omitting dependency label features.
	RGAT incorporates relational features into attention calculation and aggregation process to obtain more informative representations. 
	As shown in Figure~\ref{fig:rgat}, the RGAT calculates two attention distributions, named as node-aware attention and relation-aware attention, taking their combination as final attention weights for feature aggregation. 
	
	\noindent\textbf{Relations as input} Denoting the relation between word $w_i$ and $w_j$ as $\mathcal{R}_{ij}$, we transform $\mathcal{R}_{ij}$ into a vector $r_{ij} \in \mathbb{R}^{d_r}$, where $d_r$ is the dimension of relation embeddings. 
	During the training process, the relation embeddings are jointly optimized with the model.
	
	\noindent\textbf{Relation-aware attention}
	Inspired by previous work on SAN~\cite{shaw-etal-2018-self}, we consider relation features when calculating attentions weights between nodes.
	In practice, the RGAT method calculates an unnormalized node-aware attention $e^{N}$ together with an unnormalized relation-aware attention $e^R$ simultaneously.
	In particular, the node-aware attention $e^{N}$ of the $l$th layer is the same as Equation~\ref{eq:dotattention}:
	\begin{equation}
	e^{N}_{ij} =
	\begin{cases}
		 f(h^{l-1}_{i}, h^{l-1}_{j}), & j \in \mathcal{N}(i)\\
		-\text{inf}, & \text{otherwise}
	\end{cases}
	\label{eq:ratt1}
	\end{equation}
	while the relation-aware attention weight is given by:
	\begin{equation}
	e^{R}_{ij} =
	\begin{cases}
		f(h^{l-1}_{i}, r_{ij}), & j \in \mathcal{N}(i)\\
		-\text{inf}, & \text{otherwise}
	\end{cases}
	\label{eq:ratt2}
	\end{equation} 
	where $r_{ij}$ denotes the vector representation of relation $\mathcal{R}_{ij}$. 
	It should be noted that the relation embedding $r_{ij}$ is shared among multiple layers and attention heads.
	
	The above two types of attention scores are combined and normalized as:
	\begin{equation}
	\hat{\alpha}_{ij} = \frac{\exp(e^{N}_{ij} + e^{R}_{ij})}{\sum_{j' \in \mathcal{N}(i)} \exp{(e^{N}_{ij'} + e^{R}_{ij'})}}.
	\label{eq:ratt3}
	\end{equation}
	
	In this way, the resulting attention scores take both node features and relation features into consideration.
	
	\noindent\textbf{Relation-aware feature aggregation} Relations can also be important in the feature aggregation process, as these additional fine-grained information can be used to enrich the feature of each word $h_i^l$. 
	To this end, we use the hidden feature vector of neighbor words $h_j^{l-1}$ together with their corresponding relation vector $r_{ij}$ as inputs to update the representation of $h_i^{l-1}$:
	\begin{equation}
	h^{l}_i = \mathop{||}_{k=1}^{Z} \sigma\big(\sum_{j \in \mathcal{N}(i)} \hat{\alpha}_{ij}^{lz} (W^{lz}_Vh^{l-1}_{j} + W^l_{Vr}r_{ij})\big),
	\end{equation}
	where $W^l_{Vr} \in \mathbb{R}^{\frac{d}{Z}\times d_r}$ is a parameter matrix.
	
	In order to learn deep features, we apply a stacked relational graph attention network with multiple layers.
	
	\subsection{Pooling and Feature Fusion}
	With a contextual encoder (Section~\ref{sec:contextual}) and a RGAT encoder (Section~\ref{sec:rgat}), we obtain the contextual and syntax-aware target mention vectors, denoted as $\{h_{i}, h_{i+1},...,h_{i+m-1}\}$ and $\{\hat{h}_{i}, \hat{h}_{i+1},...,\hat{h}_{i+m-1}\}$, respectively. 
	We then apply a pooling function over these vectors to obtain two global representation $h_{con}$ and $h_{syn}$:
	\begin{equation}
	\begin{split}
	h_{con} &= pool\big({h_{i}, h_{i+1},...,h_{i+m-1}}\big), \\
	h_{syn} &= pool\big({\hat{h}_{i}, \hat{h}_{i+1},...,\hat{h}_{i+m-1}}\big). \\
	\end{split}
	\label{eq:fusion}
	\end{equation}
	
	In our implementation, $pool$ is an average pooling function\footnote{We also tried other functions such as max pooling and random pooling, without observing significant improvements.}.
	
	In order to learn a composite representation which contains both contextual and syntax features, a fine-grained feature fusion mechanism is introduced to control the fusion ratio.
	We implement the feature fusion mechanism based on the gating mechanism~\cite{HochSchm97,cho-etal-2014-learning}.
	Formally, the syntax representation $h_{syn}$ is fused into the contextual representation $h_{con}$ by:
	\begin{equation}
	h_f = g \circ h_{syn} + (1-g)\circ h_{con},
	\end{equation}
	where $\circ$ is element-wise product operation, and $g$ is a gate computed by:
	\begin{equation}
	g = \sigma(W_g[h_{syn};h_{con}]+b_g).
	\end{equation}

	In above equation, $[h_{syn};h_{con}]$ is the concatenation of $h_{syn}$ and $h_{con}$, $W_g$ and $b_g$ are model parameters.
	\subsection{The Classifier}
	The classifier is a fully connected network, which takes the fused representation $h_f$ as input and computes the probability of each sentiment class $c$:
	\begin{equation}
	P(y=c) = \frac{\exp(Wh_f+b)_c}{\sum_{c' \in \mathcal{C}}\exp(Wh_f+b)_{c'}},
	\label{eq:prob}
	\end{equation}
	where $W$ and $b$ are tunable model parameters, and $\mathcal{C}$ is the set of sentiment classes.
	
	Given a set of training instances $D=\{d_1, d_2, ..., d_N\}$, the training objective is a cross-entropy loss with $L_2$ regularization:
	\begin{equation}
	\ell = -\sum_{i=1}^{N}\sum_{c \in \mathcal{C}} I(y=c) log(P(y=c)) + \lambda \lVert \Theta \rVert^2,
	\label{eq:loss}
	\end{equation}
	where $I$ is an indicator function, $N$ is the number of training examples, $\lambda$ is a regularization hyperparameter and $\Theta$ denotes all the set of parameters in the model.
	\section{Experiments}
	\begin{table}
		\centering
		\caption{Statistics of datasets.}
		\label{tab:1}
		\begin{tabular}{lcccccc}
			\toprule
			\multirow{2}{*}{\textbf{Dataset}}& \multicolumn{2}{c}{\textbf{Positive}} &\multicolumn{2}{c}{\textbf{Negative}} & \multicolumn{2}{c}{{\textbf{Neutral}}}\\
			\cmidrule(lr){2-3} \cmidrule(lr){4-5} \cmidrule(lr){6-7}
			&Train &Test &Train &Test &Train &Test\\
			\midrule
			Restaurant &2164 &727 &807 &196 &637 &196 \\
			Laptop &976 &337 &851 &128 &455 &167 \\
			Twitter &1507 &172 &1528 &169 &3016 &336 \\
			{MAMS} &{3380} &{400} &{2764} &{329} &{5042} &{607} \\
			\bottomrule
		\end{tabular}
	\end{table}
	We conduct experiments on $4$ benchmark datasets, including the Restaurant reviews (\textit{Restaurant}) and Laptop reviews (\textit{Laptop}) datasets of SemEval 2014~\cite{pontiki-etal-2014-semeval}, the ACL14 \textit{Twitter} dataset~\cite{dong-etal-2014-adaptive} and the MAMS dataset~\cite{jiang-etal-2019-challenge}. These datasets are labeled with three sentiment polarities: \textit{positive}, \textit{neutral} and \textit{negative}. 
	The number of samples in each category are summarized in Table~\ref{tab:1}.
	\subsection{Settings}
	We adopt similar experimental settings as previous work~\cite{sun-etal-2019-aspect,huang-carley-2019-syntax}. 
	Two types of contextual encoders are considered: 1) The BiLSTM-based encoder; 2) the BERT-based encoder.
	With regard to the BiLSTM-based encoder, 300-dimensional GloVe vectors as adopted for word representation. The values are fixed during the training process.
	The dimensions of POS-tag, position and relation embeddings are set as 30.
	The dropout rate on input word embeddings is 0.7, and the $L_2$ regularization term $\lambda=10^{-5}$.
	The Adamax~\cite{Adam} optimizer with a learning rate of $10^{-3}$ is adopted to train our models.
	For the BERT-based encoder, we adopt a pre-trained BERT\footnote{We use pre-trained BERT-base-uncased model from \url{https://github.com/huggingface/transformers}.} for fine-tuning.
	The dropout rate on BERT embeddings is 0.1, and regularization term $\lambda = 10^{-5}$.
	The Adam~\cite{Adam} optimizer with a learning rate $2*10^{-5}$ is adopted for model training.
	For RGAT encoder, we use $5$ attention heads in each layer. 
	The input/output dimension of each layer is listed in Table~\ref{tab:module}. 
	We use the Deep Biaffine Parser~\cite{dozat2017deep}\footnote{\url{https://github.com/yzhangcs/parser}} to obtain dependency trees.
	
	We consider two metrics for model evaluation: Accuracy and Macro-Averaged F1. 
	The latter is more appropriate for datasets with unbalanced classes.
	Pairwise t-test is conducted on both Accuracy and Macro-Averaged F1.

	\subsection{Baselines}
	We compare our model with the state-of-the-art systems without and with syntactic knowledge.
	The syntax-free baselines include:
	\begin{itemize}
		
		\item[-] \textbf{SVM} applys a support vector machine (SVM,~\cite{Cortes95support-vectornetworks}) model over extensive features for classification.
		
		\item[-] \textbf{IAN}~\cite{ma2017-ijcai} represents the target and context interactively via two LSTMs and attention mechanism. 
		
		\item[-] \textbf{TNet}~\cite{li-etal-2018-transformation} transforms BiLSTM embeddings into target-specific embeddings, and uses a CNN for encoding. 
		
		\item[-] \textbf{MGAN}~\cite{fan-etal-2018-multi} exploits a BiLSTM to capture contextual information and a multi-grained attention mechanism to capture the relationship between aspect and context.
		
		\item[-] \textbf{AOA}~\cite{huang2018aspect} introduces an Attention-over-Attention module to model the interaction between aspects and context sentences jointly.
		
		\item[-] \textbf{AEN}~\cite{Song2019AttentionalEN} adopts attentional encoder network for feature representation and modeling semantic interactions between target and context.
		
		\item[-] {\textbf{CapsNet}~\cite{jiang-etal-2019-challenge} utilizes capsule networks~\cite{Sabour17nips} to model complicated relationships between target mentions and contexts.}
		
		\item[-] \textbf{BERT-PT}~\cite{xu-etal-2019-bert} uses a post-training approach on a pre-trained BERT model to improve the performance for review
		reading comprehension and targeted aspect sentiment classification. 
		
		\item[-] \textbf{BERT-SPC}~\cite{Song2019AttentionalEN} feeds ``[CLS]'' + sentence + ``[SEP]'' + target mention + ``[SEP]'' into a pre-trained BERT model, and then uses pooled embedding for classification.
		
		\item[-] {\textbf{CapsNet-BERT}~\cite{jiang-etal-2019-challenge} builds capsule networks on the top of BERT layers to predict sentiment polarities.}
	\end{itemize}
	
	\begin{table}
		\centering{
			\caption{\label{tab:module} Input/output dimension of model sub-modules.} 
			\scalebox{1}{
				\begin{tabular}{lcc}
					\toprule
					{\textbf{Name}}& \multicolumn{1}{c}{\textbf{GloVe}} & \multicolumn{1}{c}{\textbf{BERT}} \\
					\midrule
					Embedding layer &1/300 &1/768  \\
					Contextual encoder layer &360/100 &768/768 \\
					RGAT encoder layer &100/100 &100/100  \\
					Feature fusion layer & 200/50 & 868/868 \\
					Classifier layer &50/3 &868/3  \\
					\bottomrule
				\end{tabular}
			}
		}
	\end{table}
	
	The syntax-based baselines are: 
	\begin{itemize}
		\item[-] \textbf{AdaRNN}~\cite{dong-etal-2014-adaptive} learns the sentence representation toward target via RNN semantic composition over a dependency tree. 
		
		\item[-] \textbf{PhraseRNN}~\cite{Nguyen2015PhraseRNNPR} extends AdaRNN by adding two phrase composition functions. The PhraseRNN takes a dependency tree as well as a constituent tree as input.
		
		\item[-] \textbf{SynAttn}~\cite{he-etal-2018-effective} incorporates syntactic distance into the attention mechanism to model the interaction between target mention and context.
		
		\item[-] \textbf{CDT}~\cite{sun-etal-2019-aspect} and \textbf{ASGCN}~\cite{zhang-etal-2019-aspect} integrate dependency trees with GCN for aspect representation learning. Compared with CDT, ASGCN additionally apply attention mechanism to obtain final representation.
		
		\item[-] \textbf{TD-GAT}~\cite{huang-carley-2019-syntax} and \textbf{TD-GAT-BERT}~\cite{huang-carley-2019-syntax} apply GAT to capture the syntax structure and improves it with LSTM to model relation across layers.
	\end{itemize}
	
	For fair comparison, we leave out baselines which rely on external resources such as auxiliary sentences~\cite{sun-etal-2019-utilizing} and other domain/language review corpus~\cite{xu-etal-2019-bert,Gao19bertABSA,Yang2019AML}.
	
	\subsection{Main Results}
	\begin{table*}
		\centering{
			\caption{\label{tab:2} Performance comparison of different models on the benchmark datasets. The best performance are
				bold-typed, $^\dagger$ denotes that the model requires dependency tree as input. Results underlined indicate that the proposed method is significantly better than state-of-the-art model at significance level p$<$0.01.} 
			\scalebox{1}{
				\begin{tabular}{llcccccccc}
					\toprule
					\multirow{2}{*}{\textbf{Category}}& \multirow{2}{*}{\textbf{Model}}& \multicolumn{2}{c}{\textbf{Restaurant}} &\multicolumn{2}{c}{\textbf{Laptop}} & \multicolumn{2}{c}{{\textbf{Twitter}}} & \multicolumn{2}{c}{\textbf{{MAMS}}} \\
					\cmidrule(lr){3-4} \cmidrule(lr){5-6} \cmidrule(lr){7-8} \cmidrule(lr){9-10} 
					& &Accuracy &Macro-F1 &Accuracy &Macro-F1 &Accuracy &Macro-F1 &{Accuracy} &{Macro-F1} \\
					\midrule
					\multirow{8}{*}{\textbf{w/o Syn.}}& SVM & 80.16 &- &70.49 &- &63.40 &63.30 &{-} &{-} \\
					& IAN &78.60 &- &72.10 &- &- &- &{76.60} &{-} \\
					& TNet &80.69 &71.27 &76.54 &71.75 &${74.97}$ &73.60 &{-} &{-} \\
					& MGAN &81.25 &71.94 &75.39 &72.47 &72.54 &70.81 &{-} &{-}\\
					& AOA &81.20 &- &74.5 &- &- &- &{77.26} &{-}\\
					& AEN &80.98 &72.14 &73.51 &69.04 &72.83 &69.81 &{66.72} &{-}\\
					& CapsNet &80.79 &- &- &- &- &- &{79.78} &{-}\\
					& BERT-PT & 84.95 &76.96 &78.07 &75.08 &- &- &{-} &{-} \\
					& BERT-SPC &84.46 &76.98 &78.99 &75.03 &73.55 & 72.14 &{82.82} &{81.90} \\
					& AEN-BERT &83.12 &73.76 &79.93 &76.31 &74.71 &73.13 &{-} &{-}\\
					& CapsNet-BERT &85.93 &- &- &- &- &- &{83.39} &{-}\\
					\midrule
					\multirow{9}{*}{\textbf{w Syn.}}& AdaRNN &- &- &- &- &66.30 &65.90 &{-} &{-}\\
					& PhraseRNN &66.20 &59.32 &- &- &- &- &{-} &{-}\\
					& SynAttn &80.45 &71.26 &72.57 &69.13 &- &- &{-} &{-} \\
					& ASGCN &80.77 &72.02 &75.55 &71.05 &72.15 &70.40 &{-} &{-} \\
					& CDT &82.30 &74.02 &77.19 &72.99 &74.66 &73.66 &{80.70} &{79.79} \\
					& TD-GAT &81.20 &- &74.00 &- &- &- &{-} &{-}\\
					& TD-GAT-BERT &83.00 &- &80.10 &- &- &- &{-} &{-}\\
					\midrule
					\multirow{3}{*}{\textbf{Ours}} & Transformer &80.78 &72.10 &74.09 &69.42 &72.78 &70.23&{79.63} &{78.92}\\
					& RGAT &$\underline{83.55}$ &$\underline{75.99}$ &$\underline{78.02}$ &$\underline{74.00}$ & $\underline{75.36}$ &$\underline{74.15}$ & $\underline{{81.75}}$ &$\underline{{80.87}}$ \\
					& Ours-RGAT-BERT &\underline{\textbf{86.68}} &\underline{\textbf{80.92}} &\underline{\textbf{80.94}} &\underline{\textbf{78.20}} &\underline{\textbf{76.28}} &\underline{\textbf{75.25}} &\underline{\textbf{{84.52}}} &\underline{\textbf{{83.74}}} \\
					\bottomrule
				\end{tabular}
			}
		}
	\end{table*}
	Table~\ref{tab:2} shows the results of different models.
	We first compare RGAT with the baseline (Transformer), which replace RGAT with a Transformer network. 
	RGAT significantly ($p<0.01$) improves the performance, with an accuracy improvement of 2.85 percent on average.
	Similarly, the average F1 score increases by 2.84 percent. 
	This indicates that dependency trees can provide useful information for the targeted aspect sentiment classification task, which is consistent with observations by recent work~\cite{sun-etal-2019-aspect,huang-carley-2019-syntax}.
	It is reasonable that Transformer gives weaker results than the state-of-the-art models, as the Transformer model adopts a simple pooling function rather than attention or CNN for target-context interaction modeling.
	
	Compared with systems that do not rely on syntactic feature (SVM, IAN, TNet, MGAN, AOA, AEN, CapsNet), the RGAT model gives significantly ($p<0.01$) better results. 
	This observation is consistent with the results of our baseline.
	In addition, compared with recent work which takes dependency trees but without relation labels as input (AdaRNN, PhraseRNN, SynAttn, ASGCN, CDT, TD-GAT), 
	RGAT also gives better results across all datasets.
	In particular, RGAT outperforms the state-of-the-art CDT model by 1.25 and 1.97 points on Restaurant with regard to accuracy and F1, respectively.
	Furthermore, the pre-trained BERT model can significantly boost the performance of each approach. 
	With the help of BERT, the proposed model achieves better results than all the baselines, giving accuracy of 86.68, 80.94, 76.28 and 84.52 on Restaurant, Laptop, Twitter and MAMS, respectively. 
	To our knowledge, we obtain the best reported results on all the datasets without using external resources.
	
	\section{Analysis}
	\subsection{Ablation Study}
	\begin{figure*}
		\centering 
		\subfigure[]{\includegraphics[width=0.47\hsize]{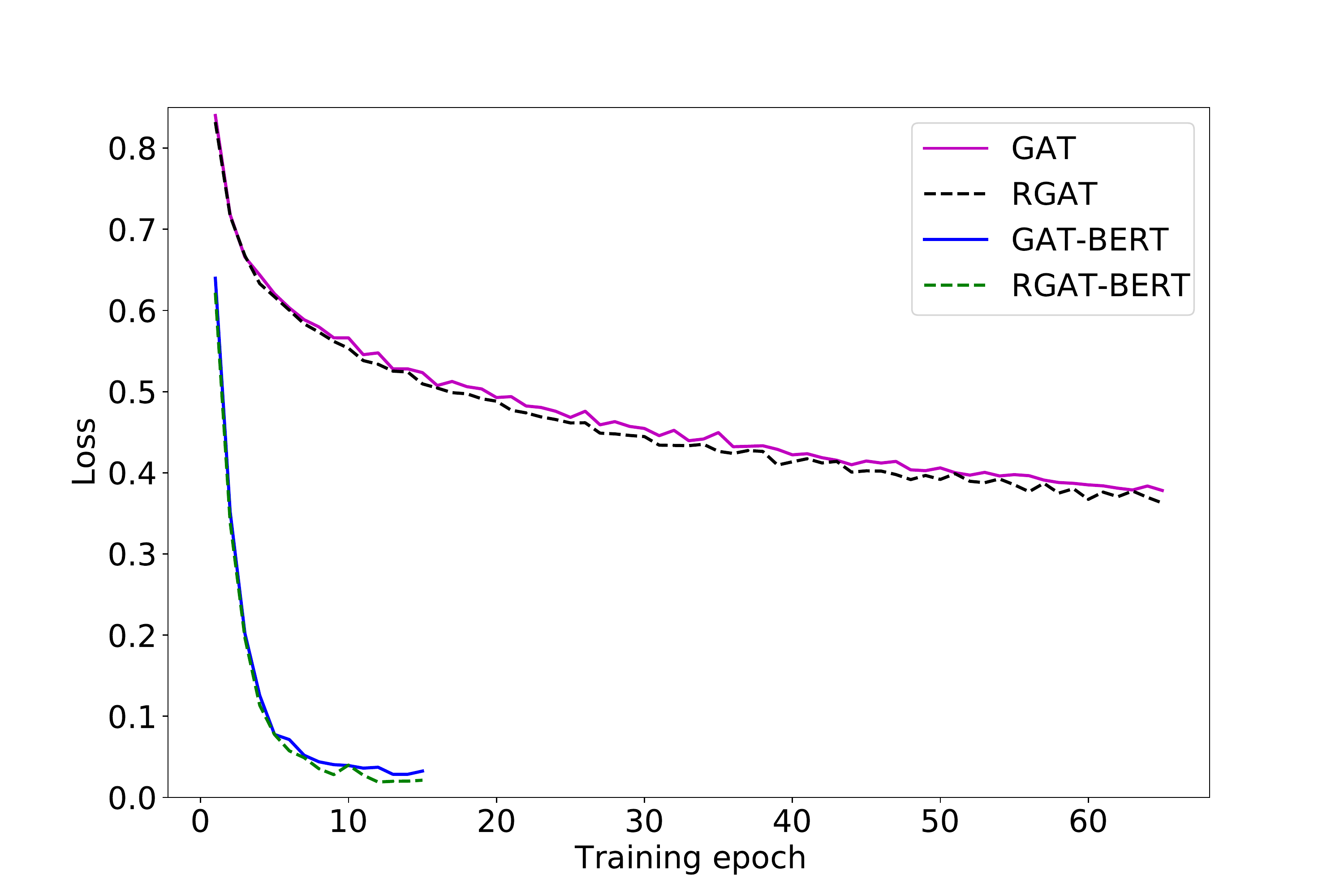}\label{fig:curve-loss}} \hspace{0.20in}
		\subfigure[]{\includegraphics[width=0.47\hsize]{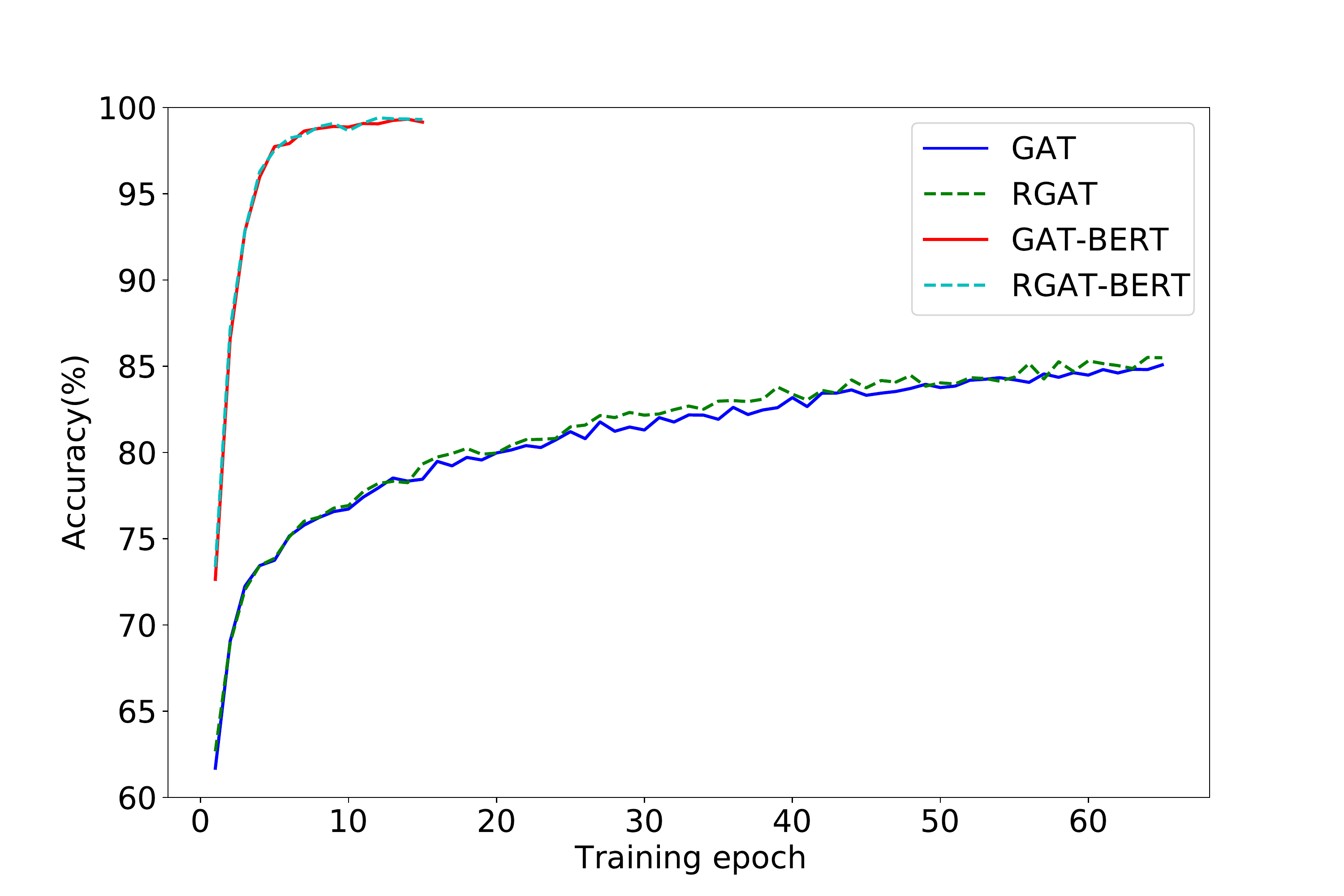}\label{fig:curve-acc}}\\ 
		\caption{The learning curve of GAT, RGAT, GAT-BERT and RGAT-BERT on MAMS.}
		\label{fig:curve}
	\end{figure*}
	We present an ablation study on the effects of structure information and relation information.
	The experiments are conducted on Restaurant, Twitter and MAMS as these three datasets have more training instances than Laptops.
	We consider three ablation baselines for comparison, including 1) Transformer: we replace the relational graph attention layers with self-attention layers. This model can be viewed as our model without dependency tree as supervision; and 2) GAT: the relation graph attention layers are substituted by graph attention layers, serving as our model without relation features; 3) {GAT-Ratt: we remove the relation-aware feature aggregation module of RGAT to test the effectiveness of the relation-aware attention independently.}
	
	\begin{table}
		\centering{
			\caption{\label{tab:3} Ablation study on the restaurant and laptop datasets. The best performance are bold-typed. Ratt refers to relational attention.} 
			\scalebox{1}{
				\begin{tabular}{lcccc}
					\toprule
					\multirow{2}{*}{\textbf{Model}}& \multicolumn{2}{c}{\textbf{Restaurant}} &\multicolumn{2}{c}{\textbf{Twitter}} \\
					\cmidrule(lr){2-3} \cmidrule(lr){4-5}
					&Accuracy &Macro-F1 &Accuracy &Macro-F1 \\
					\midrule
					Transformer &80.78 &72.10 &72.78 &70.23 \\
					GAT &82.41 &75.10 &73.75 &71.58 \\
					GAT-Ratt &82.68 &75.39 &74.60 &73.05 \\
					RGAT &${83.55}$ &${75.99}$ &${75.36}$ &${74.15}$ \\
					\midrule
					Transformer-BERT &84.89 &77.90 &73.43 &72.08 \\
					GAT-BERT &85.70 &78.80 &74.67 &73.50 \\
					GAT-Ratt-BERT &86.13 &79.61 &75.35 &74.27 \\
					RGAT-BERT &\textbf{86.68} &\textbf{80.92} &\textbf{76.28} &\textbf{75.25} \\
					\bottomrule
				\end{tabular}
			}
		}
	\end{table}
	
	\noindent\textbf{Effectiveness of structural information} As shown in Table~\ref{tab:3}, the GAT model gives consistently better accuracies and F1 scores over the Transformer model, with 1.3 and 1.2 percent average improvement of accuracy and F1 score, respectively.
	In addition, the performance of GAT-BERT is also better than that of Transformer-BERT. 
	Such results indicates that explicit syntax knowledge is helpful for targeted sentiment classification.
	A possible reason for why improvement over BERT is relatively small is that structural information is contained in the BERT representations to some extent because of contextual language modeling~\cite{jawahar-etal-2019-bert}.
	
	\noindent\textbf{Effectiveness of relation-aware attention}
	GAT-Ratt consistently shows better performance over GAT.
	Specifically, GAT-Ratt outperforms GAT by 1.0 percent accuracy and 1.5 percent F1. 
	Similar improvements are also observed for BERT based models.
	Such results indicate that the relation-aware attention mechanism is useful for targeted sentiment analysis.\\
	\noindent\textbf{Effectiveness of RGAT for typed dependencies} It can be observed that RGAT gives better results than GAT-Ratt and GAT on both datasets. 
	In particular, RGAT outperforms GAT-Ratt on Twitter by 0.76 percent and 1.1 percent with regard to accuracy and F1, respectively.
	This observation confirms our intuition that the fine-grained relation information is helpful.
	Combined with relation-aware attention and relation-aware feature aggregation, the accuracy of GAT increases by 1.1 percent and 1.6 percent on Restaurant and Twitter, respectively.
	The improvement on F1 score reaches 0.9 percent and 2.6 percent, respectively.
	Similar trends can also be observed when comparing RGAT-BERT with GAT-BERT.
	Such results demonstrate that dependency labels can bring significant improvement to the baseline model, regardless of different input embeddings and contextual models.
	
	{We present the learning curves of GAT, RGAT, GAT-BERT and RGAT-BERT in Figure~\ref{fig:curve}. 
	It can be seen that RGAT gives consistently higher accuracy and lower loss compared with GAT. 
	Similarly, the results of RGAT-BERT are better than GAT-BERT.
	Such results prove that RGAT(-BERT) is more powerful than GAT(-BERT) from another perspective.
    }
	
	\subsection{Impacts of Parsing Accuracy}
	We conduct experiments to study the effects of parsing accuracy on the classification performance.
	Specifically, we consider the following settings: 1) Transformer, which serves as a baseline model without dependency tree; 2) Random Tree, which uses a random tree as input for training\footnote{We select 10 random seeds and report the averaged results.}; 3) Random Label, which uses the gold tree structure (dependency heads) but randomly permutes the dependency labels; 4) Stanford Parser, for which dependency trees are parsed by Stanford Transition-based Parser~\cite{chen-manning-2014-fast}; 5)  Biaffine Parser, which applies a Deep Biaffine Parser~\cite{dozat2017deep} to obtain dependency trees. 6) BERT Biaffine Parser, which adopts the BERT based Biaffine Parser to obtain a dependency tree.
	\begin{table}
		\centering
		\caption{Performance of different parsers. UAS refers to unlabeled attachment score and LAS refers to labeled attachment score}
		\label{tab:parser}
		\begin{tabular}{lcc}
			\toprule
			Model &UAS &LAS \\
			\midrule
			Stanford Parser &94.10 &91.49 \\
			Biaffine Parser &95.90 &94.25 \\
			BERT Biaffine Parser &\textbf{96.64} &\textbf{95.11} \\
			\bottomrule
		\end{tabular}
	\end{table}	
	The performances of the Stanford Parser and Deep Biaffine Parser on the Penn Treebank are given in Table~\ref{tab:parser}.
	\begin{table}
		\centering{
			\caption{\label{tab:parsing} The accuracy and F1 when using different dependency trees as input.}
			\scalebox{0.9}{
				\begin{tabular}{lcccc}
					\toprule
					\multirow{2}{*}{\textbf{Model}}& \multicolumn{2}{c}{\textbf{Restaurant}} &\multicolumn{2}{c}{\textbf{Twitter}} \\
					\cmidrule(lr){2-3} \cmidrule(lr){4-5}
					&Accuracy &Macro-F1 &Accuracy &Macro-F1 \\
					\midrule
					Transformer &80.78 &72.10 &72.78 &70.23 \\
					Random Tree &78.97 &70.46 &71.20 &68.75 \\
					Random Labels &82.03 &74.67 &73.63 &71.42 \\
					Stanford Parser &83.37 &75.82 &75.13 &74.12 \\
					Biaffine Parser &83.55 &\textbf{75.99} &\textbf{75.36} &\textbf{74.15} \\
					BERT Biaffine Parser &\textbf{83.64} &75.91 &75.28 &74.10 \\
					\bottomrule
				\end{tabular}
			}
		}
	\end{table}
	\begin{figure}[t]
		\centering\includegraphics[width=0.48\textwidth]{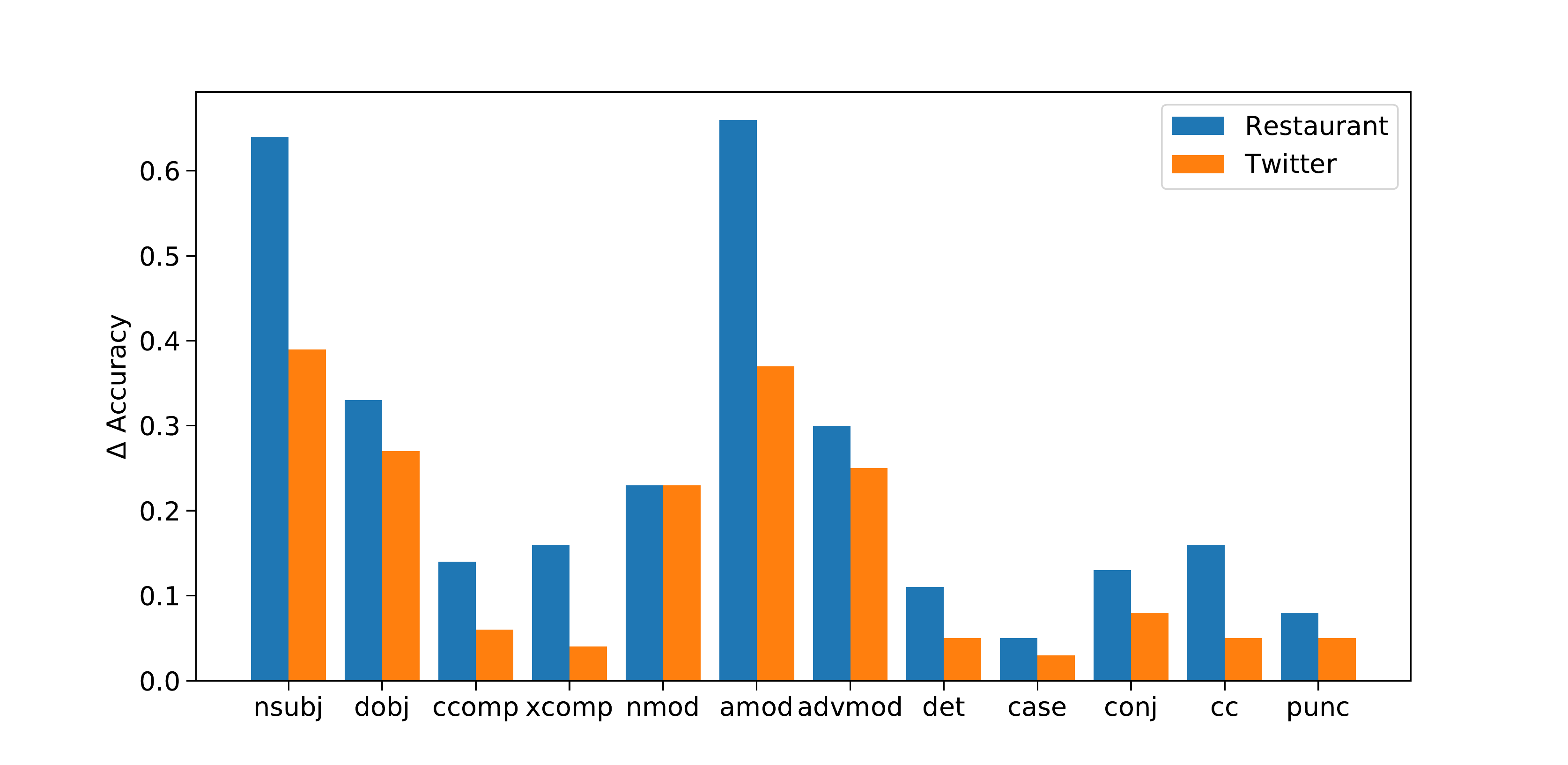}
		\footnotesize\caption{The decrement of accuracy (compared with the baseline) on Restaurant and Twitter when removing different dependency labels.}
		\label{fig:dep}
	\end{figure}
	
	\begin{table*}[!t]
		\centering{
			{\caption{\label{tab:cofficient} Model performance on benchmark datasets.}
				\scalebox{1}{
					\begin{tabular}{lcccccc}
						\toprule
						\multirow{2}{*}{\textbf{Model}}& \multicolumn{2}{c}{\textbf{Restaurant}} &\multicolumn{2}{c}{\textbf{Twitter}} &\multicolumn{2}{c}{\textbf{MAMS}}\\
						\cmidrule(lr){2-3} \cmidrule(lr){4-5} \cmidrule(lr){6-7}
						&Accuracy &Macro-F1 &Accuracy &Macro-F1 &Accuracy &Macro-F1\\
						\midrule
						Transformer &80.78 &72.10 &72.78 &70.23 &79.63 &78.92 \\
						RGAT-\textit{weighted factors} &83.01 &{76.35} &{75.51} &73.84 &81.52 &80.68 \\
						RGAT &{83.55} &{75.99} &75.36 &{74.15} &{81.75} &{80.87} \\
						\midrule
						Transformer-BERT &84.89 &77.90 &73.43 &72.08 &82.23 &81.47 \\
						RGAT-BERT-\textit{weighted factors} &86.23 &79.83 &75.84 &75.13  &{84.30} &{83.50} \\
						RGAT-BERT&\textbf{86.68} &\textbf{80.92} &\textbf{76.28} &\textbf{75.25} &\textbf{84.52} &\textbf{83.74} \\
						\bottomrule
					\end{tabular}
				}
		}}
	\end{table*}
	\begin{table}
		\centering{
			{\caption{\label{tab:cofficient-value} The values of trainable parameter $\beta_1$ and $\beta_2$. We report mean$^{\pm std}$ value of multiple RGAT layers.} 
				\scalebox{1}{
					\begin{tabular}{lcccc}
						\toprule
						\multirow{2}{*}{\textbf{Dataset}}& \multicolumn{2}{c}{\textbf{RGAT}} &\multicolumn{2}{c}{\textbf{RGAT-BERT}} \\
						\cmidrule(lr){2-3} \cmidrule(lr){4-5}
						&$\beta_1$ &$\beta_2$ &$\beta_1$ &$\beta_2$ \\
						\midrule
						Restaurant &0.992$^{\pm0.19}$ &1.194$^{\pm0.17}$ &1.000$^{\pm 5e^{-5}}$ &1.000$^{\pm3e-5}$ \\
						Twitter &0.973$^{\pm0.04}$ &1.080$^{\pm0.05}$ &1.000$^{\pm 2e^{-4}}$ &1.000$^{\pm 2e^{-4}}$ \\
						MAMS &0.925$^{\pm0.15}$ &1.093$^{\pm0.22}$ &0.993$^{\pm5e-4}$ &1.000$^{\pm3e-4}$ \\
						\bottomrule
					\end{tabular}
				}
		}}
	\end{table}
	
	Table~\ref{tab:parsing} gives the results of different systems on Restaurant and Twitter.
	It can be seen that a random tree has negative influences on the classification performance.
	The Random Label model gives better results than Transformer, showing that the tree structure is still useful for targeted sentiment classification, despite label noise.
	In addition, the trees parsed by pre-trained models (Stanford, Biaffine and BERT Biaffine parser) have more positive impact on model performance than random trees.
	This is likely because pre-trained parsers can give more accurate predictions and the parsed results are more consistent.
	
	Comparing the results of different parsers, it can be seen that both Biaffine Parser and BERT Biaffine Parser give better results than the Stanford Parser, and the performance of the Biaffine Parser and the BERT Biaffine Parser are comparable. 
	Such results indicate that the classification performance is positively correlated with the parsing accuracy.
	
	\subsection{Effects of different dependency labels}
	We select the most frequent dependency labels and study their contribution to the model performance.
	In particular, the following categories of dependency labels are considered: 1) Clausal Argument Relations, including $\{$\textit{nsubj, dobj, ccomp, xcomp}$\}$; 2) Nominal Modifier Relations, including $\{$\textit{nmod, amod, advmod, det, case}$\}$; 3) Other Notable Relations, including $\{$\textit{conj, cc, punc}$\}$.
	
	Figure~\ref{fig:dep} gives the results of the RGAT model on the two benchmarks.
	We report the decrement of classification accuracy when removing different dependency labels independently.
	The accuracy of RGAT decreases most rapidly when removing \textit{nsubj} and \textit{amod} relations, indicating that \textit{nsubj} and \textit{amod} carry most important information  for classification. 
	This is intuitive as the \textit{nsubj} label represents the nominal subject relation, which is especially informative when one side of this relation is a target mention.
	Similarly, the \textit{amod} label denotes the adjective modifier relation, and is always related to the adjective which conveys the sentiment polarity.
	In addition, the \textit{dobj}, \textit{nmod} and \textit{advmod} labels are also important factors to the model performance. The accuracy decreases by a range from 0.23 to 0.32 when these labels are removed. 
	Last but not least, although having a high frequency, the \textit{det}, \textit{case} and \textit{punc} dependency labels are the most irrelevant to the classification accuracy, with accuracy decline of about 0.07, 0.04 and 0.05, respectively.
	This indicates that the label frequency is not strongly correlated to the classification accuracy, and the relation label matters.
	
	\subsection{{Node attention VS Edge attention}}

	{In order to compare the importance of node attention and edge attention, we assign two trainable weighted factors to node attention and edge attention, respectively. In particular, we modified Equation~\ref{eq:ratt3} as:
		\begin{equation}
			\hat{\alpha}_{ij} = \frac{\exp(\beta_1 e^{N}_{ij} + \beta_2 e^{R}_{ij})}{\sum_{j' \in \mathcal{N}(i)} \exp{(\beta_1 e^{N}_{ij'} + \beta_2 e^{R}_{ij'})}},
			\label{eq:ratt4}
		\end{equation}
	where $\beta_1, \beta_2 \in \mathbb{R}^1$ are model parameters.
	}

	{
	We conduct evaluation on the Restaurant and Twitter datasets, and results are shown in Table~\ref{tab:cofficient}. RGAT-\textit{weighted factors} achieves comparable performance to RGAT, and RGAT-BERT-\textit{weighted factors} gives slightly lower results compared with RGAT-BERT. 
	}
	
	{
	The values of $\beta_1$ and $\beta_2$ are shown in Table~\ref{tab:cofficient-value}. Interestingly, both values are close to 1, and this phenomenon is more obvious when using the BERT model. Such results can explain why the performance of RGAT-\textit{weighted factors} is comparable  to RGAT, and further indicate that node attention is of equal importance to relation attention.
	}
	\subsection{Impacts of Model Depth}
	\begin{figure}[!t]
		\centering\includegraphics[width=0.45\textwidth]{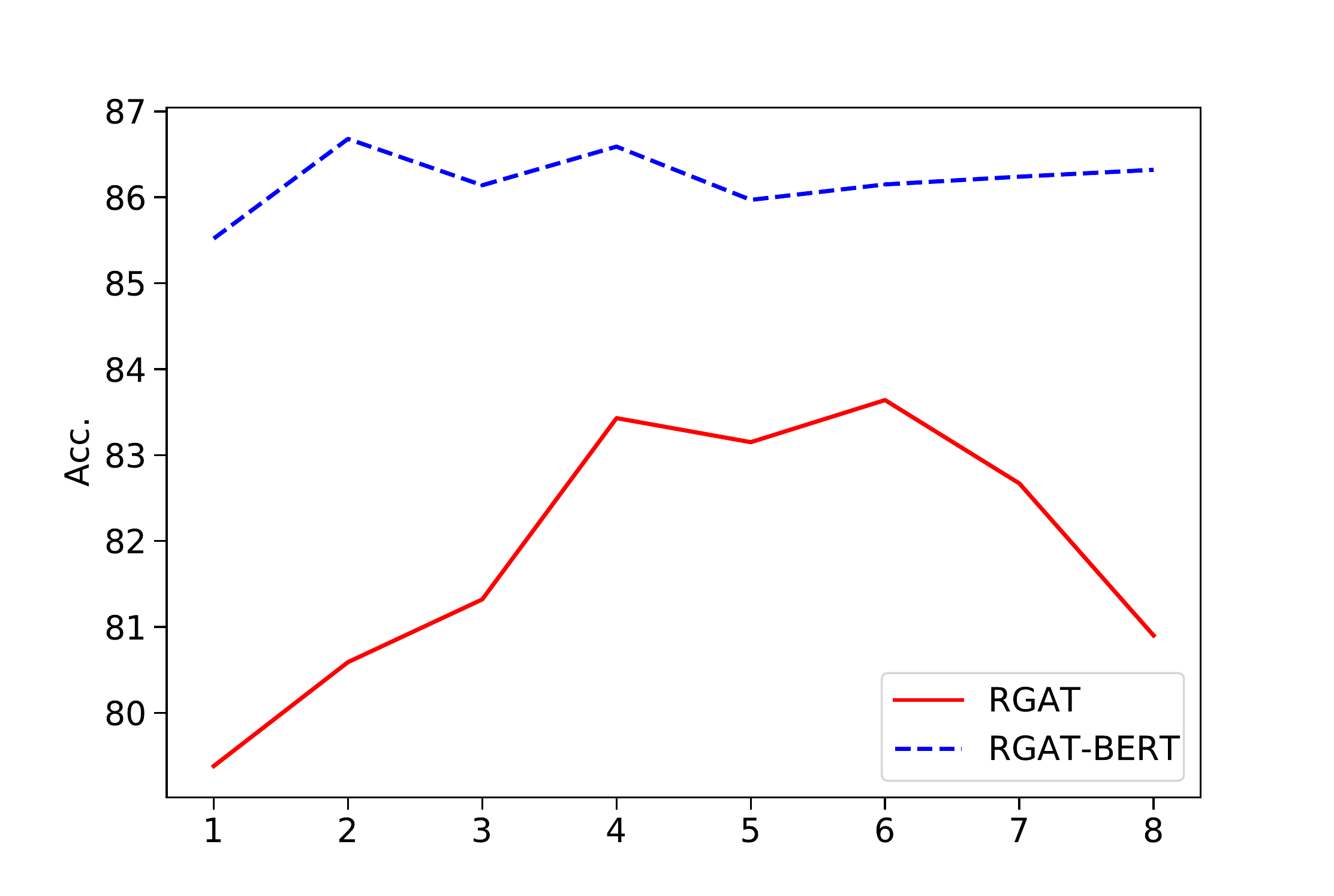}
		\footnotesize\caption{The effects of model depth on classification accuracy on Restaurant.}
		\label{fig:03}
	\end{figure}
	Figure~\ref{fig:03} shows the accuracy curves for RGAT and RGAT-BERT on Restaurant. 
	Different numbers of layers ranging from 1 to 8 are considered.
	For RGAT, the initial accuracy is low and then increases along with the depth, reaching the best score of $83.55$ with $6$ layers.
	This is intuitive as the target-related sentiment words can be many-hops away from the target mention, which means that multiple layers of node communication is necessary for passing information from relevant context to the target mention.
	In contrast to RGAT, RGAT-BERT reaches the best accuracy (86.68) faster with $2$ layers, which can be because that implicate syntax contained in BERT allows faster information propagation.
	\begin{figure}[!t]
		\centering 
		\subfigure[]{\includegraphics[width=0.95\hsize,height=0.15\hsize]{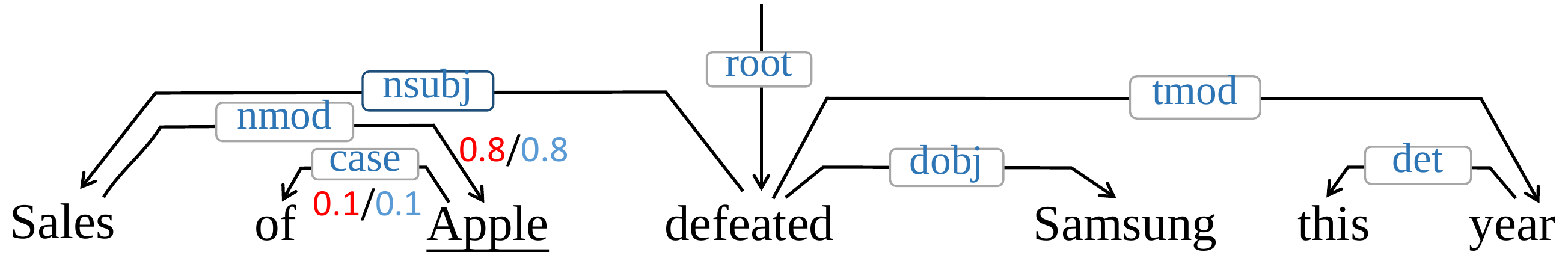}\label{fig:4-1}}\\ \vspace{-5pt}        
		\subfigure[]{\includegraphics[width=0.95\hsize]{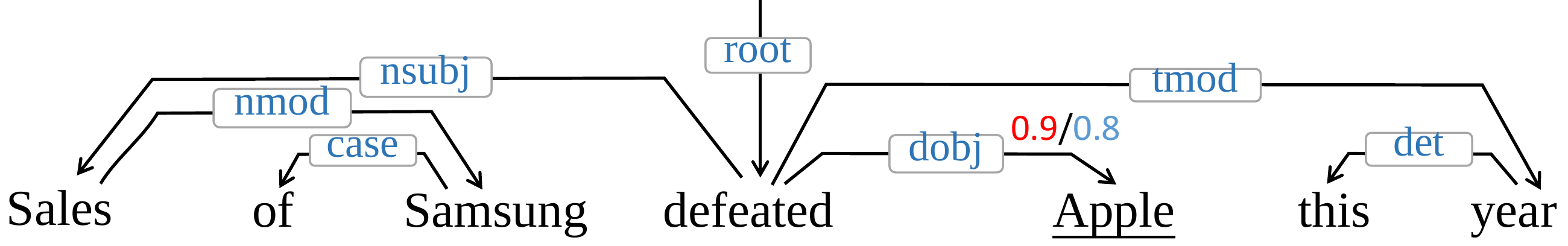}\label{fig:4-2}}\\ \vspace{-5pt} 
		\subfigure[]{\includegraphics[width=0.95\hsize]{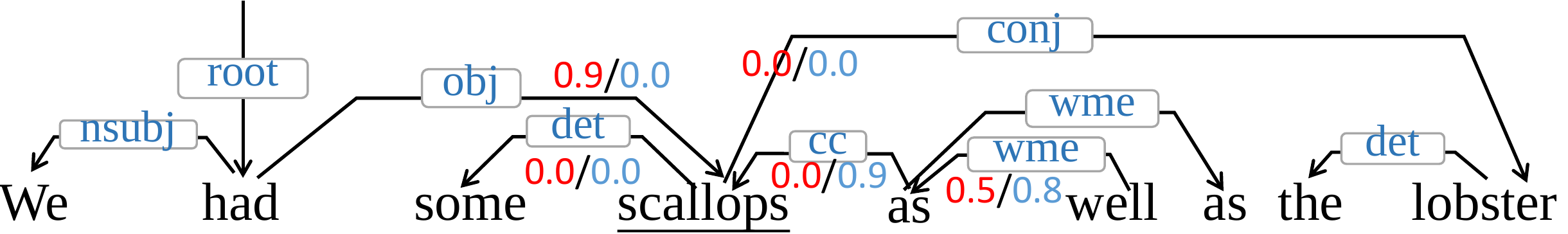}\label{fig:4-3}}\\ \vspace{-5pt} 
		\subfigure[]{\includegraphics[width=0.95\hsize]{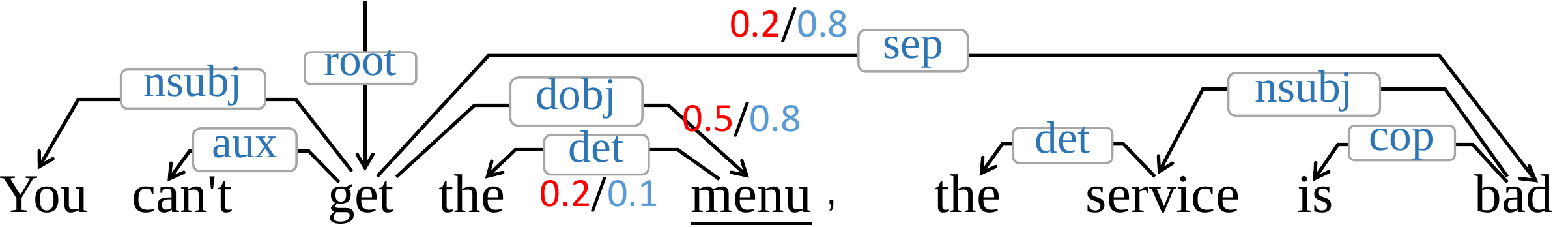}\label{fig:4-4}}\\ 
		\caption{Several samples, the numbers denote the attention weights given by RGAT (red) and GAT (blue). The target mentions are underlined.	
		}
		\label{fig:example2}
	\end{figure}
	\subsection{Case Study}
	\label{sec:case}
	We provide $4$ examples to further analyze the proposed model. 
	The attention scores over dependency edges are given for comparison.\footnote{For brevity, we omit the attention weight on self-loop edges.} 
	Consider the case shown in Figure~\ref{fig:4-1} and Figure~\ref{fig:4-2}. GAT gives a right sentiment in case 1 and a incorrect sentiment in case 2, while RGAT obtains the correct sentiment in both cases, which indicates that RGAT can differentiate similar structures, thanks to the use of dependency labels.
	In Figure~\ref{fig:4-3}, GAT predicts a ``positive'' sentiment for target mention ``scallops'', and the attention weight on the edge ``as''$\longrightarrow$``scallops'' is high, which demonstrates that GAT is negatively influenced by ``well'', which does not have positive meaning in this case. 
	In contrast, RGAT gives the correct result and the attention weight on the edge  ``as''$ \stackrel{cc}{\longrightarrow}$``scallops'' is close to zero.
	Figure~\ref{fig:4-4} shows another case where the decision of GAT is influenced by an irrelevant  word ``bad''. GAT's attention weight on the edge  ``get''$ \stackrel{cc}{\longrightarrow}$``menu'' is $0.8$ while that of RGAT is $0.5$.
	The last two examples indicate that the dependency labels have a positive influence on modeling the relationship between two words.
	\section{Discussion}
	\begin{table}
		\centering{
			\caption{\label{tab:param} Comparison of model parameters, 1M$=1e^6$.} 
			\scalebox{1}{
				\begin{tabular}{lcccc}
					\toprule
					\multirow{2}{*}{\textbf{Dataset}}& \multicolumn{2}{c}{\textbf{GloVe}} &\multicolumn{2}{c}{\textbf{BERT}} \\
					\cmidrule(lr){2-3} \cmidrule(lr){4-5}
					&BiLSTM &Ours &BERT-SPC &Ours \\
					\midrule
					Restaurant &1.53M &1.90M &109.59M &110.00M \\
					Laptop &1.25M &1.49M &109.58M &110.00M \\
					Twitter &4.13M &4.50M &110.37M &113.97M \\
					MAMS &2.71M &2.96M &109.59M &110.00M \\
					\bottomrule
				\end{tabular}
			}
		}
	\end{table}
	{
	Our approach is more complicated than previous syntax-free work, due to the modeling of typed dependencies. However, the total number of parameters in RGAT does not increase significantly. The main reason is that we use a small hidden size (100) in RGAT encoder. Table~\ref{tab:param} shows the number of parameters of our method and two syntax-free systems (BiLSTM~\cite{tang-etal-2016-effective}, BERT-SPC~\cite{Song2019AttentionalEN}).
	}

	{
	In addition, our model requires more calculations than previous work. However, the encoding process of sentence and dependency tree are independent, so that the forward calculation and backward calculation of RGAT encoder and contextual encoder can be parallelized theoretically.
	}
	
	\section{Conclusion}
	We investigated the usage of typed dependency structures for targeted sentiment classification, by extending a graph attention network encoder with relation features. Extensive experiments on four standard benchmarks show that label information is useful for sentiment classification, and our relational GAT model can effectively encode such features. Our final model gives results that are better than the existing best results in the literature.
	We further study the impact of parsing performance, dependency labels and network depth on model performance, finding that different dependency arc labels do have different effects on targeted sentiment signal propagation, thereby motivating the use of arc label information. In addition, contextualized embeddings are complementary to structured dependencies for improving the results.
	
	\ifCLASSOPTIONcaptionsoff
	\newpage
	\fi

	
	
	\bibliographystyle{IEEEtran}
	\bibliography{taslp}

\begin{thebibliography}{10}
\providecommand{\url}[1]{#1}
\csname url@samestyle\endcsname
\providecommand{\newblock}{\relax}
\providecommand{\bibinfo}[2]{#2}
\providecommand{\BIBentrySTDinterwordspacing}{\spaceskip=0pt\relax}
\providecommand{\BIBentryALTinterwordstretchfactor}{4}
\providecommand{\BIBentryALTinterwordspacing}{\spaceskip=\fontdimen2\font plus
\BIBentryALTinterwordstretchfactor\fontdimen3\font minus
  \fontdimen4\font\relax}
\providecommand{\BIBforeignlanguage}[2]{{%
\expandafter\ifx\csname l@#1\endcsname\relax
\typeout{** WARNING: IEEEtran.bst: No hyphenation pattern has been}%
\typeout{** loaded for the language `#1'. Using the pattern for}%
\typeout{** the default language instead.}%
\else
\language=\csname l@#1\endcsname
\fi
#2}}
\providecommand{\BIBdecl}{\relax}
\BIBdecl

\bibitem{jiang-etal-2011-target}
L.~Jiang, M.~Yu, M.~Zhou, X.~Liu, and T.~Zhao, ``Target-dependent twitter
  sentiment classification,'' in \emph{ACL}, 2011, pp. 151--160.

\bibitem{dong-etal-2014-adaptive}
L.~Dong, F.~Wei, C.~Tan, D.~Tang, M.~Zhou, and K.~Xu, ``Adaptive recursive
  neural network for target-dependent twitter sentiment classification,'' in
  \emph{ACL}, 2014.

\bibitem{Vo2015TargetDependentTS}
D.-T. Vo and Y.~Zhang, ``Target-dependent twitter sentiment classification with
  rich automatic features,'' in \emph{IJCAI}, 2015.

\bibitem{Zhang2016GatedNN}
M.~Zhang, Y.~Zhang, and D.-T. Vo, ``Gated neural networks for targeted
  sentiment analysis,'' in \emph{AAAI}, 2016.

\bibitem{wang-etal-2018-target}
S.~Wang, S.~Mazumder, B.~Liu, M.~Zhou, and Y.~Chang, ``Target-sensitive memory
  networks for aspect sentiment classification,'' in \emph{ACL}, 2018.

\bibitem{pang-etal-2002-thumbs}
B.~Pang, L.~Lee, and S.~Vaithyanathan, ``Thumbs up? sentiment classification
  using machine learning techniques,'' in \emph{EMNLP}, 2002.

\bibitem{Meena07}
A.~Meena and T.~V. Prabhakar, ``Sentence level sentiment analysis in the
  presence of conjuncts using linguistic analysis,'' in \emph{Advances in
  Information Retrieval}, G.~Amati, C.~Carpineto, and G.~Romano, Eds.\hskip 1em
  plus 0.5em minus 0.4em\relax Berlin, Heidelberg: Springer Berlin Heidelberg,
  2007, pp. 573--580.

\bibitem{yessenalina-etal-2010-multi}
A.~Yessenalina, Y.~Yue, and C.~Cardie, ``Multi-level structured models for
  document-level sentiment classification,'' in \emph{EMNLP}, 2010.

\bibitem{wang-etal-2016-attention}
Y.~Wang, M.~Huang, X.~Zhu, and L.~Zhao, ``Attention-based {LSTM} for
  aspect-level sentiment classification,'' in \emph{EMNLP}, 2016.

\bibitem{Wang2018TargetSensitiveMN}
S.~Wang, S.~Mazumder, B.~Liu, M.~Zhou, and Y.~Chang, ``Target-sensitive memory
  networks for aspect sentiment classification,'' in \emph{ACL}, 2018.

\bibitem{tang-etal-2016-effective}
D.~Tang, B.~Qin, X.~Feng, and T.~Liu, ``Effective {LSTM}s for target-dependent
  sentiment classification,'' in \emph{COLING}, 2016.

\bibitem{li-etal-2018-transformation}
X.~Li, L.~Bing, W.~Lam, and B.~Shi, ``Transformation networks for
  target-oriented sentiment classification,'' in \emph{ACL}, 2018.

\bibitem{huang-carley-2018-parameterized}
B.~Huang and K.~Carley, ``Parameterized convolutional neural networks for
  aspect level sentiment classification,'' in \emph{EMNLP}, 2018.

\bibitem{Ma2017InteractiveAN}
D.~Ma, S.~Li, X.~Zhang, and H.~Wang, ``Interactive attention networks for
  aspect-level sentiment classification,'' in \emph{IJCAI}, 2017.

\bibitem{tang-etal-2019-progressive}
J.~Tang, Z.~Lu, J.~Su, Y.~Ge, L.~Song, L.~Sun, and J.~Luo, ``Progressive
  self-supervised attention learning for aspect-level sentiment analysis,'' in
  \emph{ACL}, 2019.

\bibitem{Tang2016AspectLS}
D.~Tang, B.~Qin, and T.~Liu, ``Aspect level sentiment classification with deep
  memory network,'' in \emph{EMNLP}, 2016.

\bibitem{Chen2017RecurrentAN}
P.~Chen, Z.~Sun, L.~Bing, and W.~Yang, ``Recurrent attention network on memory
  for aspect sentiment analysis,'' in \emph{EMNLP}, 2017.

\bibitem{tai-etal-2015-improved}
K.~S. Tai, R.~Socher, and C.~D. Manning, ``Improved semantic representations
  from tree-structured long short-term memory networks,'' in \emph{ACL}, 2015.

\bibitem{kipf2017semi}
T.~N. Kipf and M.~Welling, ``Semi-supervised classification with graph
  convolutional networks,'' in \emph{International Conference on Learning
  Representations}, 2017.

\bibitem{velickovic2018gat}
P.~Veličković, G.~Cucurull, A.~Casanova, A.~Romero, P.~Liò, and Y.~Bengio,
  ``Graph attention networks,'' in \emph{International Conference on Learning
  Representations}, 2018.

\bibitem{sun-etal-2019-aspect}
K.~Sun, R.~Zhang, S.~Mensah, Y.~Mao, and X.~Liu, ``Aspect-level sentiment
  analysis via convolution over dependency tree,'' in \emph{EMNLP-IJCNLP},
  2019.

\bibitem{huang-carley-2019-syntax}
B.~Huang and K.~Carley, ``Syntax-aware aspect level sentiment classification
  with graph attention networks,'' in \emph{EMNLP-IJCNLP}, 2019.

\bibitem{zhang-etal-2019-aspect}
C.~Zhang, Q.~Li, and D.~Song, ``Aspect-based sentiment classification with
  aspect-specific graph convolutional networks,'' in \emph{EMNLP-IJCNLP}, 2019.

\bibitem{HochSchm97}
S.~Hochreiter and J.~Schmidhuber, ``Long short-term memory,'' \emph{NC}, 1997.

\bibitem{xue-li-2018-aspect}
W.~Xue and T.~Li, ``Aspect based sentiment analysis with gated convolutional
  networks,'' in \emph{ACL}, 2018.

\bibitem{Bahdanau2015NeuralMT}
D.~Bahdanau, K.~Cho, and Y.~Bengio, ``Neural machine translation by jointly
  learning to align and translate,'' in \emph{ICLR}, 2015.

\bibitem{Vaswani2017AttentionIA}
A.~Vaswani, N.~Shazeer, N.~Parmar, J.~Uszkoreit, L.~Jones, A.~N. Gomez,
  L.~Kaiser, and I.~Polosukhin, ``Attention is all you need,'' in \emph{NIPS},
  2017.

\bibitem{li-etal-2018-hierarchical}
L.~Li, Y.~Liu, and A.~Zhou, ``Hierarchical attention based position-aware
  network for aspect-level sentiment analysis,'' in \emph{Conll}, 2018.

\bibitem{devlin-etal-2019-bert}
J.~Devlin, M.-W. Chang, K.~Lee, and K.~Toutanova, ``{BERT}: Pre-training of
  deep bidirectional transformers for language understanding,'' in
  \emph{NAACL}, 2019.

\bibitem{Radford2018ImprovingLU}
A.~Radford, ``Improving language understanding by generative pre-training,''
  2018.

\bibitem{Lan2020ALBERTAL}
Z.~Lan, M.~Chen, S.~Goodman, K.~Gimpel, P.~Sharma, and R.~Soricut, ``Albert: A
  lite bert for self-supervised learning of language representations,'' in
  \emph{International Conference on Learning Representations}, 2020.

\bibitem{Clark2020ELECTRA}
K.~Clark, M.-T. Luong, Q.~V. Le, and C.~D. Manning, ``Electra: Pre-training
  text encoders as discriminators rather than generators,'' in
  \emph{International Conference on Learning Representations}, 2020.

\bibitem{Song2019AttentionalEN}
Y.~Song, J.~Wang, T.~Jiang, Z.~Liu, and Y.~Rao, ``Attentional encoder network
  for targeted sentiment classification,'' \emph{CoRR}, vol. abs/1902.09314,
  2019.

\bibitem{Gao19bertABSA}
Z.~{Gao}, A.~{Feng}, X.~{Song}, and X.~{Wu}, ``Target-dependent sentiment
  classification with bert,'' \emph{IEEE Access}, vol.~7, pp.
  154\,290--154\,299, 2019.

\bibitem{sun-etal-2019-utilizing}
C.~Sun, L.~Huang, and X.~Qiu, ``Utilizing {BERT} for aspect-based sentiment
  analysis via constructing auxiliary sentence,'' in \emph{NAACL}, 2019.

\bibitem{xu-etal-2019-bert}
H.~Xu, B.~Liu, L.~Shu, and P.~Yu, ``{BERT} post-training for review reading
  comprehension and aspect-based sentiment analysis,'' in \emph{NAACL}, 2019.

\bibitem{li-etal-2019-exploiting}
X.~Li, L.~Bing, W.~Zhang, and W.~Lam, ``Exploiting {BERT} for end-to-end
  aspect-based sentiment analysis,'' in \emph{Proceedings of the 5th Workshop
  on Noisy User-generated Text (W-NUT 2019)}, 2019.

\bibitem{Goldberg2019AssessingBS}
Y.~Goldberg, ``Assessing bert's syntactic abilities,'' \emph{CoRR}, vol.
  abs/1901.05287, 2019.

\bibitem{clark2019what}
K.~Clark, U.~Khandelwal, O.~Levy, and C.~D. Manning, ``What does bert look at?
  an analysis of bert's attention,'' in \emph{BlackBoxNLP@ACL}, 2019.

\bibitem{Qiu2011OpinionWE}
G.~Qiu, B.~Liu, J.~Bu, and C.~Chen, ``Opinion word expansion and target
  extraction through double propagation,'' \emph{Computational Linguistics},
  vol.~37, pp. 9--27, 2011.

\bibitem{Liu2013OpinionTE}
K.~Liu, H.~Xu, Y.~Liu, and J.~Zhao, ``Opinion target extraction using
  partially-supervised word alignment model,'' in \emph{IJCAI}, 2013.

\bibitem{Nguyen2015PhraseRNNPR}
T.~H. Nguyen and K.~Shirai, ``Phrasernn: Phrase recursive neural network for
  aspect-based sentiment analysis,'' in \emph{EMNLP}, 2015.

\bibitem{he-etal-2018-effective}
R.~He, W.~S. Lee, H.~T. Ng, and D.~Dahlmeier, ``Effective attention modeling
  for aspect-level sentiment classification,'' in \emph{Proceedings of the 27th
  International Conference on Computational Linguistics}, 2018.

\bibitem{wang-etal-2020-relational}
\BIBentryALTinterwordspacing
K.~Wang, W.~Shen, Y.~Yang, X.~Quan, and R.~Wang, ``Relational graph attention
  network for aspect-based sentiment analysis,'' in \emph{Proceedings of the
  58th Annual Meeting of the Association for Computational Linguistics}.\hskip
  1em plus 0.5em minus 0.4em\relax Online: Association for Computational
  Linguistics, Jul. 2020, pp. 3229--3238. [Online]. Available:
  \url{https://www.aclweb.org/anthology/2020.acl-main.295}
\BIBentrySTDinterwordspacing

\bibitem{shaw-etal-2018-self}
P.~Shaw, J.~Uszkoreit, and A.~Vaswani, ``Self-attention with relative position
  representations,'' in \emph{NAACL}, 2018.

\bibitem{Sundermeyer2012LSTMNN}
M.~Sundermeyer, R.~Schl{\"u}ter, and H.~Ney, ``Lstm neural networks for
  language modeling,'' in \emph{INTERSPEECH}, 2012.

\bibitem{google16}
Y.~Wu, M.~Schuster, Z.~Chen, Q.~V. Le, M.~Norouzi, W.~Macherey, M.~Krikun,
  Y.~Cao, Q.~Gao, K.~Macherey, J.~Klingner, A.~Shah, M.~Johnson, X.~Liu,
  L.~Kaiser, S.~Gouws, Y.~Kato, T.~Kudo, H.~Kazawa, K.~Stevens, G.~Kurian,
  N.~Patil, W.~Wang, C.~Young, J.~Smith, J.~Riesa, A.~Rudnick, O.~Vinyals,
  G.~Corrado, M.~Hughes, and J.~Dean, ``Google's neural machine translation
  system: Bridging the gap between human and machine translation,''
  \emph{CoRR}, vol. abs/1609.08144, 2016.

\bibitem{song-etal-2018-graph}
L.~Song, Y.~Zhang, Z.~Wang, and D.~Gildea, ``A graph-to-sequence model for
  {AMR}-to-text generation,'' in \emph{ACL}, 2018.

\bibitem{cho-etal-2014-learning}
K.~Cho, B.~van Merri{\"e}nboer, C.~Gulcehre, D.~Bahdanau, F.~Bougares,
  H.~Schwenk, and Y.~Bengio, ``Learning phrase representations using {RNN}
  encoder{--}decoder for statistical machine translation,'' in \emph{EMNLP},
  2014.

\bibitem{pontiki-etal-2014-semeval}
M.~Pontiki, D.~Galanis, J.~Pavlopoulos, H.~Papageorgiou, I.~Androutsopoulos,
  and S.~Manandhar, ``{S}em{E}val-2014 task 4: Aspect based sentiment
  analysis,'' in \emph{SemEval}, 2014.

\bibitem{jiang-etal-2019-challenge}
\BIBentryALTinterwordspacing
Q.~Jiang, L.~Chen, R.~Xu, X.~Ao, and M.~Yang, ``A challenge dataset and
  effective models for aspect-based sentiment analysis,'' in \emph{Proceedings
  of the 2019 Conference on Empirical Methods in Natural Language Processing
  and the 9th International Joint Conference on Natural Language Processing
  (EMNLP-IJCNLP)}.\hskip 1em plus 0.5em minus 0.4em\relax Hong Kong, China:
  Association for Computational Linguistics, Nov. 2019, pp. 6280--6285.
  [Online]. Available: \url{https://www.aclweb.org/anthology/D19-1654}
\BIBentrySTDinterwordspacing

\bibitem{Adam}
D.~P. Kingma and J.~Ba, ``Adam: A method for stochastic optimization,'' 2015.

\bibitem{dozat2017deep}
T.~Dozat and D.~C. Manning, ``Deep biaffine attention for neural dependency
  parsing,'' \emph{international conference on learning representations}, 2017.

\bibitem{Cortes95support-vectornetworks}
C.~Cortes and V.~Vapnik, ``Support-vector networks,'' in \emph{Machine
  Learning}, 1995, pp. 273--297.

\bibitem{ma2017-ijcai}
D.~Ma, S.~Li, X.~Zhang, and H.~Wang, ``Interactive attention networks for
  aspect-level sentiment classification,'' in \emph{IJCAI}, 2017.

\bibitem{fan-etal-2018-multi}
F.~Fan, Y.~Feng, and D.~Zhao, ``Multi-grained attention network for
  aspect-level sentiment classification,'' in \emph{EMNLP}, 2018.

\bibitem{huang2018aspect}
B.~Huang, Y.~Ou, and K.~M. Carley, ``Aspect level sentiment classification with
  attention-over-attention neural networks,'' in \emph{International Conference
  on Social Computing, Behavioral-Cultural Modeling and Prediction and Behavior
  Representation in Modeling and Simulation}.\hskip 1em plus 0.5em minus
  0.4em\relax Springer, 2018, pp. 197--206.

\bibitem{Sabour17nips}
\BIBentryALTinterwordspacing
S.~Sabour, N.~Frosst, and G.~E. Hinton, ``Dynamic routing between capsules,''
  in \emph{Advances in Neural Information Processing Systems}, I.~Guyon, U.~V.
  Luxburg, S.~Bengio, H.~Wallach, R.~Fergus, S.~Vishwanathan, and R.~Garnett,
  Eds., vol.~30.\hskip 1em plus 0.5em minus 0.4em\relax Curran Associates,
  Inc., 2017, pp. 3856--3866. [Online]. Available:
  \url{https://proceedings.neurips.cc/paper/2017/file/2cad8fa47bbef282badbb8de5374b894-Paper.pdf}
\BIBentrySTDinterwordspacing

\bibitem{Yang2019AML}
H.~Yang, B.~Zeng, J.~Yang, Y.~Song, and R.~Xu, ``A multi-task learning model
  for chinese-oriented aspect polarity classification and aspect term
  extraction,'' \emph{Neurocomputing}, vol. 419, pp. 344--356, 2021.

\bibitem{jawahar-etal-2019-bert}
G.~Jawahar, B.~Sagot, and D.~Seddah, ``What does {BERT} learn about the
  structure of language?'' in \emph{ACL}, 2019.

\bibitem{chen-manning-2014-fast}
D.~Chen and C.~Manning, ``A fast and accurate dependency parser using neural
  networks,'' in \emph{EMNLP}, 2014.

\end{thebibliography}
\end{document}